\documentclass[lettersize,journal]{IEEEtran}
\usepackage{amsmath,amsfonts}
\usepackage{algorithmic}
\usepackage{algorithm}
\usepackage{array}
\usepackage[caption=false,font=normalsize,labelfont=sf,textfont=sf]{subfig}
\usepackage{amssymb}
\usepackage{textcomp}
\usepackage{stfloats}
\usepackage{url}
\usepackage{verbatim}
\usepackage{graphicx}
\usepackage{cite}
\usepackage{booktabs}

\begin{document}

\title{Otters++: A Time-to-first-spike Based Energy Efficient Optical Spiking Transformer}

\author{Zhanglu Yan$^{1 }$, Jiayi Mao$^{2}$, Kaiwen Tang$^{1}$,  Fanfan Li$^{2}$,  Gang Pan$^4$, Tao Luo$^5$,  \text{Bowen Zhu$^2$, Qianhui Liu$^{1,3}$, Weng-Fai Wong$^1$} \thanks{
$^1$Department of Computer Science, National University of Singapore, $^2$School of Engineering, Westlake University, $^3$School of Artificial Intelligence, Shandong University, $^4$College of Computer Science and Technology, Zhejiang University, $^5$Agency for Science, Research and Technology, Singapore. %Corresponding to: Qianhui Liu, Jiayi Mao
} \\
%\texttt{\{zlyan,qhliu, wongwf\}@nus.edu.sg} 
}

\markboth{Journal of \LaTeX\ Class Files,~Vol.~14, No.~8, August~2021}%
{Shell \MakeLowercase{\textit{et al.}}: A Sample Article Using IEEEtran.cls for IEEE Journals}


\maketitle

\begin{abstract}
Spiking neural networks (SNNs) are promising for energy-efficient inference, and time-to-first-spike (TTFS) coding is especially attractive because each neuron fires at most once. In practice, however, this benefit is often reduced by the cost of computing a temporal decay term and multiplying it by the synaptic weight. We address this issue by turning a physical hardware “bug,” the natural signal decay in optoelectronic devices, into the main computation of TTFS, named Otters++. Specifically, we use the measured decay of a custom In$_2$O$_3$ optoelectronic synapse to directly realize the TTFS temporal term, removing the need for explicit digital decay computation. To scale this idea to Transformer models, we establish a layer-wise functional equivalence between the Otters++ and a quantized neural network (QNN), and develop a hybrid training method that uses device-faithful SNN computation in the forward pass and QNN straight-through gradients through the equivalent QNN path in the backward pass, together with model distillation. This avoids differentiation through discrete first-spike events and reduces the over-sparsity problem in direct TTFS-SNN training. We further make training aware of measured device noise by sampling run-to-run variation, and refine the system-level energy model by accounting for device sharing and multi-hop communication. On GLUE dataset, Otters++ improves the average score to 84.17\% while maintaining a clear energy advantage over prior spiking Transformer baselines. These results show that physically grounded TTFS computing can be efficient, trainable, and robust under realistic hardware effects.
\end{abstract}


\section{Introduction}
\label{sec:intro}

Large language models(LLMs) have achieved remarkable success across language tasks, but their computational and energy costs remain a major obstacle to deployment on edge and resource-constrained platforms~\cite{lin2023pushing, jegham2025hungry}. This challenge has motivated increasing interest in spiking neural networks (SNNs), whose sparse and event-driven computation offers a potential path toward low-power inference~\cite{xing2024spikellm, tangsorbet,liu2024lite,liu2025human}. Among different neural coding schemes, time-to-first-spike (TTFS) is especially appealing because each neuron fires at most once within a coding window, thereby maximizing temporal sparsity and, in principle, reducing spike traffic and data movement~\cite{yu2023ttfs,zhaottfsformer,yan2024reconsidering}. However, the practical efficiency of TTFS is often overstated. In conventional TTFS implementations, the arrival time of each spike must still be converted into a numerical value through a temporal decay function, such as an exponential or linear decay, and this value is then multiplied by the synaptic weight. As a result, the apparent benefit of sparse spike coding is partly offset by additional function evaluation, multiplication, and memory access~\cite{wei2023temporal,che2024ettfs}. This hidden overhead raises a fundamental question: \textbf{Can TTFS preserve its sparsity advantage without paying the cost of digital temporal decoding?}

%In this work, we answer this question by moving the temporal computation from digital logic into device physics. Instead of treating the natural signal decay of an optoelectronic device as an undesirable non-ideality~\cite{alqahtani2025light}, we use it as the temporal computation required by TTFS. 

In this work, we answer this question by moving the temporal computation from digital logic into device physics. Instead of treating the natural signal decay of an optoelectronic device as an undesirable non-ideality~\cite{alqahtani2025light}, we use it as the temporal computation required by TTFS. Specifically, the measured decay response of a custom In$_2$O$_3$ optoelectronic synapse directly realizes the TTFS temporal term, removing the need to explicitly evaluate the decay function in software or digital hardware. In this way, Otters++ repurposes a physical hardware ``bug'' as a computational primitive, fusing temporal modulation and synaptic computation into the same physical process.

%Specifically, the measured decay response of a custom In$_2$O$_3$ optoelectronic synapse is directly used to implement the TTFS temporal term, so that the system no longer needs to explicitly evaluate the decay function in software or digital hardware. In this way, a physical hardware ``bug'' is repurposed as a computational primitive, fusing temporal modulation and synaptic computation into the same physical process. Thus, \textbf{Our method, Otters++, uses the natural decay of this device's optical signal to perform the required computation.} This approach integrates storage and computation into a single physical step, thereby addressing the overhead of traditional TTFS.

While our optical hardware mitigates the computational overhead of TTFS, a second major barrier remains: the inherent difficulty of training such networks, especially for complex architectures such as transformers. Direct optimization of TTFS-SNNs is challenging because discrete first-spike events and quantized firing times make gradient propagation fragile, particularly under sparse or missing spikes. This leads to unstable training and the over-sparsity problem commonly observed in direct SNN optimization~\cite{wei2023temporal}. 
In our setting, this training difficulty is further amplified by hardware-induced mismatch and variation~\cite{li2024artificial}. Physical sampling error arises when the ideal firing time falls between realizable sampling points of the measured device response, while run-to-run device fluctuations perturb the decay curve and reduce robustness. If these effects are considered only after training, they introduce a mismatch between the trained model and the deployed Otters++ path. 

To address these challenges, we establish a layer-wise functional equivalence between the Otters++ and an unsigned QNN, and use this equivalence to develop a hybrid SNN-forward/QNN-backward training framework. The forward pass preserves the hardware-faithful Otters++ computation. At each discretized spike time, the device response is sampled from the interval between the measured upper and lower response curves, allowing the model to account for realistic hardware uncertainty during training. The backward pass avoids direct differentiation through sparse and discontinuous first-spike events, and instead computes gradients through the functionally equivalent nominal QNN with the straight-through estimator. As a result, the proposed training scheme maintains physical fidelity in the forward computation, enables stable optimization, and improves robustness under measured device variation.

%A second practical challenge is robustness under realistic hardware variation. The device response is not perfectly fixed, but shows run-to-run fluctuations in the measured decay behavior~\cite{li2024artificial}. Instead of modeling this effect with only synthetic perturbations, we incorporate the measured variation envelope directly into training. In the forward pass, the device response at each discretized spike time is sampled from the interval between the measured upper and lower curves. In the backward pass, gradients are still computed through the equivalent nominal QNN path, which keeps optimization efficient while minimizing the expected loss under realistic device uncertainty. In this way, the training process is not only device-aware, but also explicitly robust to the variation observed in hardware measurements. To further stabilize training, we also adopt knowledge distillation from a pretrained teacher. Specifically, the student is guided by both prediction-level and hidden-state supervision, which improves optimization under quantization and device noise. This complements the proposed SNN-forward/QNN-backward training scheme and helps preserve accuracy under realistic hardware conditions.

Beyond training, practical deployment also requires a system-level energy evaluation.
Prior studies of SNN efficiency often focus on operation counts, while underemphasizing data movement and weight access, which can dominate energy in real deployments. We therefore develop an energy model that explicitly captures these costs and adapt it to the physical constraints of Otters++. Further, because analog-read energy depends on how optoelectronic devices are allocated and shared across neurons, we incorporate device sharing into the energy model. We also extend the analysis to multi-hop communication, so that the evaluated energy reflects different communication distances. These refinements provide a more realistic assessment of the energy benefits of physically grounded TTFS computing.
Together, these components enable a physically grounded TTFS framework that combines optical temporal computation, stable SNN training, hardware-variation robustness, and system-level energy evaluation.
Evaluated on GLUE, Otters++ improves the average score by 0.95 points over the original Otters baseline and by 3.34--4.37 points over other SNN-based Transformer baselines. Under the same hardware model, it reduces per-layer energy by 1.84$\times$ over SpikingLM, 3.02$\times$ over Sorbet, and up to 5.68$\times$ over SpikingBERT~\cite{tangsorbet,xingspikelm,bal2024spikingbert}.

%On GLUE, Otters++ improves the average score by 0.95 points over the original Otters baseline and by 3.34–4.37 points over other SNN-based Transformer baselines. Further, Otters++ reduces per-layer energy by 1.84× over SpikingLM and 3.02× over Sorbet, and by up to 5.68× over SpikingBERT under the same hardware model.~\cite{tangsorbet,xingspikelm,bal2024spikingbert}. 

%Taken together, these ideas form a physically grounded TTFS framework that addresses computation, trainability, robustness, and deployment realism.
%Evaluated on the GLUE benchmark, Otters++ achieves average accuracy of 3\% higher than previous leading SNNs while demonstrating a 1.77-3.04$\times$ improvement in energy efficiency compared to baselines like Sorbet and SpikingLM~\cite{tangsorbet,xingspikelm}. Notably, this energy-efficiency gain is validated by a rigorous and comprehensive analysis that goes beyond the simplistic metrics commonly used in prior SNN research. While previous work often only counted compute operations (e.g., additions vs. multiplications), our analysis provides a more realistic estimate. It is grounded in measurements from a commercial 22nm process and provides a full accounting of compute, data movement, and memory access costs, making our efficiency claims robust. 

The main contributions of this work are threefold:

\begin{enumerate}

    \item We introduce a physically grounded TTFS computing mechanism that turns the natural optical decay of optoelectronic devices into the core temporal computation. Rather than treating decay as a hardware nonideality, Otters++ exploits it to replace the costly digital temporal evaluation required by conventional TTFS-SNNs.

      \item We develop a hybrid \emph{SNN-forward/QNN-backward} training framework for Otters++. Based on a layer-wise functional equivalence between the Otters++ TTFS-SNN and QNN, the forward pass preserves the device-faithful TTFS computation, while the backward pass uses STE-based gradients through the equivalent QNN path. We further incorporate measured run-to-run device variation into the forward pass, enabling robust training under realistic hardware uncertainty.

    \item We refine the energy evaluation of Otters++ with a system-level model that accounts for compute, data movement, memory access, analog-read cost, device sharing, and multi-hop communication, providing a more realistic estimate of energy efficiency.
\end{enumerate}

\section{Background}
 \subsection{Optoelectronic synapse}
 
 An optoelectronic synapse is a neuromorphic device that emulates biological synaptic functions by using optical signals to modulate its electrical conductance. These devices are renowned for their potential for extreme energy efficiency, broader bandwidth and faster signal transmission in neuromorphic computing, which are key advantages over purely electronic counterparts~\cite{xie2024emerging,wang2023advanced}. Recent studies have reported energy consumption reaching the femtojoule (fJ)/spike level, comparable to that of biological synapses and substantially lower than that of conventional CMOS neuron devices~\cite{shi2022fully, wang2024monolithic}. Among various implementations, oxide thin-film transistors (TFT) are regarded as viable candidates for optoelectronic synapses due to their low leakage current and capability for large-area, flexible fabrication. Solution-based fabrication further offers the advantages of low cost, simplified processing, and facile compositional control. Previous reports have shown that solution-processed devices exhibit uniform performance, operational stability, and low energy consumption~\cite{li2025double}.  Building upon these advances, this work employs the mature and reliable oxide-TFT platform to develop the Otters++ spiking neuron.

 \subsection{Time-to-first-spike SNN}
In contrast to rate-based encoding, which uses the frequency of spikes to represent information, TTFS encoding leverages the precise timing of a single spike. The core principle is that a stronger input stimulus causes a neuron's membrane potential to rise faster, reaching its firing threshold sooner. Thus, the information is encoded in the arrival time of the first—and only—spike within a given time window, $T$. This approach maximizes temporal sparsity and is highly efficient, as each neuron fires at most once~\cite{che2024efficiently}.

The operation of a standard TTFS neuron involves two phases. First, the neuron integrates incoming spikes, updating its membrane potential $V_j^l(t)$. Second, it compares this potential to a firing threshold $\theta^l(t)$. A spike is generated at the first time step $t$ where the potential meets or exceeds the threshold:
\begin{equation}
s_j^l(t) = 
\begin{cases}
1, & \text{if } V_j^l(t) \ge \theta^l(t) \\
0, & \text{otherwise}
\end{cases}
\end{equation}

However, the asynchronous nature of SNNs, combined with the ``fire-as-early-as-possible'' objective of TTFS, can lead to another problem. If a presynaptic neuron fires after a postsynaptic neuron has already fired,  its spike becomes invalid for membrane potential accummalation. To solve this, we employ a Dynamic Firing Threshold (DFT) model that enforces a synchronous, layer-by-layer processing schedule~\cite{wei2023temporal}. The threshold for any neuron in layer $l$ is set to infinity outside of a designated time window, effectively ensuring that layer $l$ is only active from time $T \cdot l$ to $T \cdot (l+1)$:
\begin{equation}
\theta^l(t) = 
\begin{cases}
\theta^l_{\text{dynamic}}(t), & \text{if } T \cdot l \le t \le T \cdot (l+1) \\
+\infty, & \text{otherwise}
\end{cases}
\end{equation}
This scheduling guarantees that all spikes from a preceding layer are processed before the current layer can fire, thus preserving the valid causal relationship.

\subsection{Energy Calculation}
\label{subsec:energy_calculation}

To enable a fair comparison across full-precision Transformers, quantized Transformers, and SNN-based models, we adopt a unified energy accounting framework~\cite{yan2024reconsidering}. 
Specifically, we decompose the energy of each layer into two dominant parts: 1) the energy of the linear projections, denoted by $E_{\text{fc}}$, and 2) the energy of the attention score computation, denoted by $E_{\text{score}}$. For both parts, the total energy includes arithmetic cost, memory-access cost, data-movement cost, and state-related overhead such as leakage, threshold comparison, and output write-back. We consider a spatial dataflow architecture where information (e.g., spike packets) is communicated over a Network-on-Chip (NoC)~\cite{yan2024reconsidering}. This architecture is representative of modern specialized hardware such as neuromorphic chips like Loihi~\cite{8589981} and dataflow AI accelerators like Tenstorrent~\cite{tenstorrent} and Sambanova~\cite{prabhakar2022sambanova}. We consider the control logic energy to be negligible as our analysis focuses on specialized accelerator designs where such overhead is minimal~\cite{yan2024reconsidering}.

Unless otherwise stated, $B$ denotes the batch size, $S$ the sequence length, $C_i$ and $C_o$ the input and output channel dimensions of a linear layer, $h$ the number of attention heads, and $d_k$ the per-head key/query dimension. For full-precision and quantized models, $\gamma \in [0,1]$ denotes the effective activation density. For SNNs, $T$ denotes the number of time steps, $s_r \in [0,1]$ denotes the average spike rate. The hardware primitives used in the model are denoted as follows: $E_{\text{MAC}}$ for one multiply-accumulate operation, $E_{\text{ACC}}$ for one accumulate operation, $E_{\text{CMP}}$ for one comparison, $E_{\text{SUB}}$ for one subtraction, $E_{\text{clamp}}$ for one clamping operation, $E_{\text{leakage}}$ for one unit-time leakage cost, $E^{\text{Read}}_{\text{weight}}$ and $E^{\text{Read}}_{\text{kv}}$ for reading weights and key/value operands, $E^{\text{Write}}_{\text{kv}}$ for writing the generated key/value outputs, and $E^{\text{sparse}}_{\text{move}}$ for moving one bit of sparse data.

\noindent\textbf{Full-Precision BERT (FP32).}
For the full-precision baseline, each valid input activation contributes a standard MAC together with one weight read and the corresponding data movement. Since the activations are represented in FP32, the sparse movement cost scales with 32 bits. In addition, each output incurs leakage, clamping, and key/value write-back overhead. The resulting energy is modeled as
\begin{align}
E_{\text{fc}} = &\; B \cdot S \cdot C_o \cdot \Bigl[ C_i \cdot E_{\text{leakage}} + 2E_{\text{clamp}} + E^{\text{Write}}_{\text{kv}} \nonumber \\
&\qquad\qquad + \gamma \cdot C_i \cdot \bigl( E_{\text{MAC}} + E^{\text{Read}}_{\text{weight}} + 32 \cdot E^{\text{sparse}}_{\text{move}} \bigr) \Bigr], \\[6pt]
E_{\text{score}} = &\; B \cdot h \cdot S^2 \cdot \Bigl[ d_k \cdot E_{\text{leakage}} + 2E_{\text{clamp}} \nonumber \\
&\qquad\qquad + d_k \cdot \gamma \cdot \bigl( E^{\text{Read}}_{\text{kv}} + E_{\text{MAC}} + 32 \cdot E^{\text{sparse}}_{\text{move}} \bigr) \Bigr].
\end{align}

The term $B \cdot S \cdot C_o$ counts the total number of output neurons in the projection layer, while $B \cdot h \cdot S^2$ counts the total number of pairwise attention-score evaluations across all heads. %Compared with the other settings, the FP32 model has the largest movement cost because each transmitted activation uses 32 bits.

\noindent\textbf{Quantized BERT.}
For the quantized baseline, the structure of the energy model remains the same, but the representation width is reduced from 32 bits to $\log_2(T+1)$ bits, corresponding to the quantized activation/state space with $T+1$ discrete levels. Therefore, the dominant reduction comes from the data-movement term, while leakage, clamping, and write-back overhead remain unchanged under the same architectural schedule. The corresponding energy is
\begin{align}
E_{\text{fc}} = & B \cdot S \cdot C_o \cdot \Bigl[ C_i \cdot E_{\text{leakage}} + 2E_{\text{clamp}} + E^{\text{write}}_{kv} \nonumber \\
& + \gamma \cdot C_i \cdot \bigl( E_{\text{MAC}} + E^{\text{Read}}_{\text{weight}} + \log_2(T+1)E^{\text{sparse}}_{\text{move}} \bigr) \Bigr] \\[6pt]
E_{\text{score}} = & B \cdot h \cdot S^2 \cdot \Bigl[ d_k \cdot E_{\text{leakage}} + 2E_{\text{clamp}} \nonumber \\
& + d_k \cdot \gamma \cdot \bigl( E^{\text{Read}}_{\text{kv}} + E_{\text{MAC}} + \log_2(T+1) \cdot E^{\text{sparse}}_{\text{move}} \bigr) \Bigr]
\end{align}

%This formulation highlights that quantization mainly reduces the communication cost of token representations, while still preserving the dense MAC-based computation pattern of conventional Transformers.

\noindent\textbf{Typical SNNs.}
For conventional SNNs, computation unfolds over $T$ time steps. Each neuron now incurs time-dependent state maintenance cost, including membrane leakage and threshold comparison at every step. In addition, the synaptic update is event-driven: only active spikes participate in accumulation, weight access, and sparse data movement, which are scaled by both the spike rate $s_r$ and the simulation length $T$. The energy is modeled as
\begin{align}
E_{\text{fc}}= & B \cdot S \cdot C_o \cdot \Bigl[ C_i \cdot T \cdot E_{\text{leakage}} + T \cdot \bigl(E_{\text{CMP}} + s_r \cdot E_{\text{SUB}}\bigr)  \nonumber \\
& + E^{\text{Write}}_{\text{kv}}+C_i \cdot s_r \cdot T \cdot \bigl( E_{\text{ACC}} + E^{\text{Read}}_{\text{weight}} + E^{\text{sparse}}_{\text{move}} \bigr) \Bigr] \\[10pt]
E_{\text{score}}= & B \cdot h \cdot S^2 \cdot \Bigl[ d_k \cdot T \cdot E_{\text{leakage}} + T \cdot \bigl(E_{\text{CMP}} + s_r \cdot E_{\text{SUB}}\bigr) \nonumber \\
& + d_k \cdot s_r \cdot T \cdot \bigl( E^{\text{Read}}_{\text{kv}} + E_{\text{ACC}} + E^{\text{sparse}}_{\text{move}} \bigr) \Bigr]
\end{align}

\section{Methods}
\label{sec:methods}

This section presents the Otters++ framework from device construction to training and system-level evaluation. We first introduce the optoelectronic neuron and the Transformer architecture built on it. We then show how Otters++ can be constructed as a layer-wise equivalent form of a QNN, which enables a hybrid \emph{SNN-forward/QNN-backward} training method. Next, we extend this training scheme to measured device variation through noise-aware forward sampling. Finally, we present the energy model used to assess deployment efficiency.

 \begin{figure*}[htbp]
    \centering
    \includegraphics[width=0.9\linewidth]{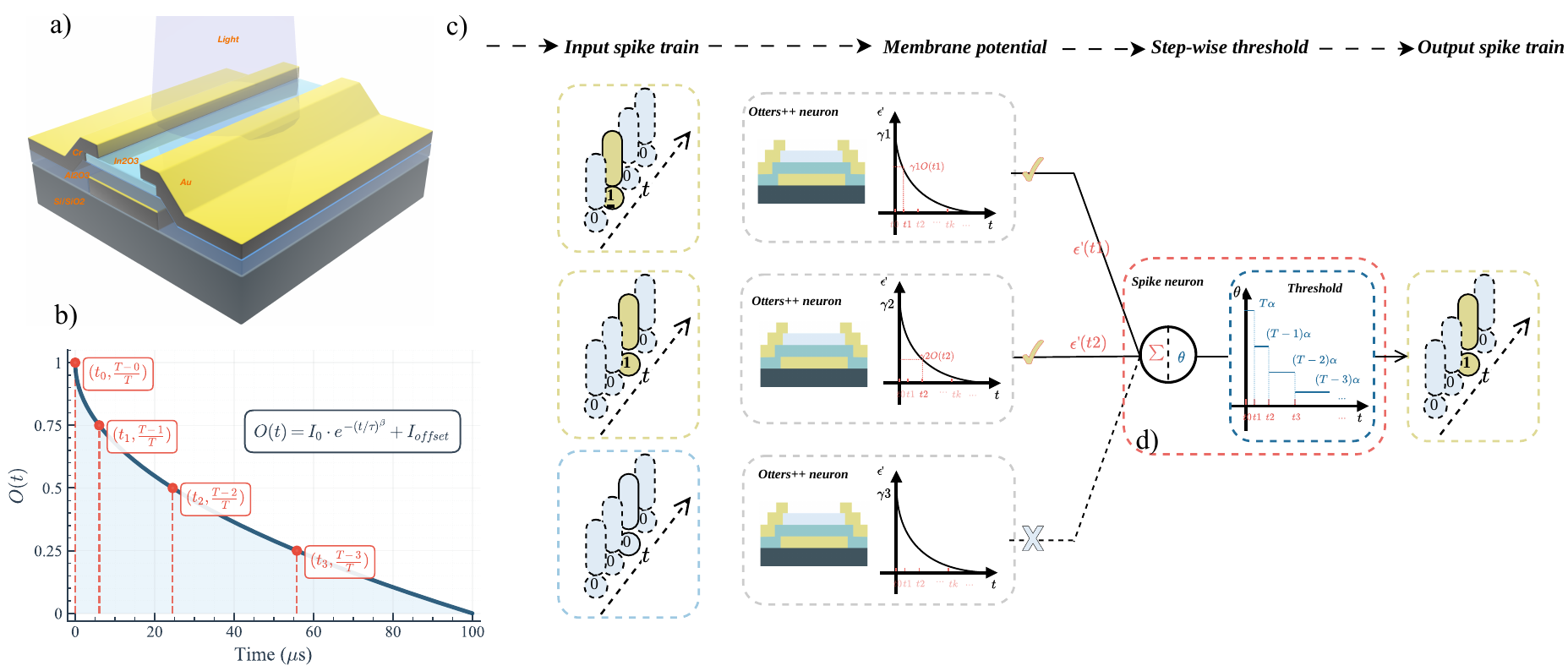}
\caption[Device and workflow]{Device and workflow: (a) the custom-fabricated In$_2$O$_3$ thin-film transistor (TFT); (b) measured decay curve of the device response; (c) Otters++ neuron workflow.}

    \label{fig:otters}
\end{figure*}

\subsection{Otters++ Spiking Neuron}
\label{subsec:Otters++_neuron}

The core of Otters++ is an optoelectronic synapse that physically realizes the temporal modulation required by time-to-first-spike (TTFS) computation. Each synapse consists of two parts: a custom-fabricated In$_2$O$_3$ thin-film transistor (TFT), which provides the physical decay response, and an analog-to-digital converter (ADC), which maps the analog response to a digital post-synaptic signal. Device fabrication details are given in Section~IV-A.

Under a fixed light intensity, the TFT exhibits a stable nonlinear decay curve~\cite{li2024active,liang2022printable}, which is modeled as
\begin{equation}
O(t) = I_0 \exp\!\left(-\left(\frac{t}{\tau}\right)^{\beta}\right) + I_{\text{offset}},
\end{equation}

This decay forms the temporal component of the post-synaptic potential (PSP). However, the device nonlinearity introduces a key challenge. For the QNN--SNN equivalence to hold, spike timing must correspond to uniformly spaced logical values. In particular, a spike at time $t_k$ should encode the quantized value $(T-k)/T$. Since the device decay $O(t)$ is nonlinear, the physical times $\{t_k\}$ at which the device output matches these target values are generally non-uniform. Therefore, a naive design based on a fixed threshold and uniform time sampling cannot satisfy the required equivalence.

To reconcile the nonlinear device physics with the linear quantization grid, we use a dynamic step-wise threshold under a uniform physical clock. Specifically, we precompute a set of physical sampling times $\{t_k\}_{k=0}^{T-1}$ from the inverse device response, and allow the neuron to fire only at these discrete times. The threshold changes only at $\{t_k\}$, so that the first firing event directly encodes the intended quantized level. The full construction is formalized in Section~\ref{subsec:qnn_equiv}.

The ADC applies a synapse-dependent scaling factor $\gamma_{ij}^l$ to the device response. The PSP generated by a presynaptic spike is
\begin{equation}
\epsilon'(t) = \gamma_{ij}^{l}\, O(t).
\label{eq:psp}
\end{equation}
The membrane potential of neuron $j$ in layer $l$ accumulates these PSPs:
\begin{equation}
V_j^{l}(t) = V_j^{l}(t-1) + \sum_{i:\, s_i^{l-1}(t)=1} \epsilon'(t).
\label{eq:membrane_update}
\end{equation}
A neuron fires when its membrane potential first reaches the dynamic threshold:
\begin{equation}
t^{\,l}_{\text{spike},j} = \min \left\{ t_k \,\middle|\, V_j^{l}(t_k) \ge \theta^{l}(t_k) \right\}.
\label{eq:spike_time}
\end{equation}
Following the TTFS paradigm, the neuron is deactivated after firing, so that each neuron emits at most one spike per inference cycle.

\subsection{Otters++ Transformer Architecture}
\label{subsec:Otters++_arch}

A main challenge in building a spiking Transformer is the matrix multiplication in self-attention, i.e., $QK^\top$. In rate-coded SNNs, one operand can often be treated as a binary spike train, which turns multiplication into selective accumulation. This simplification does not directly apply to TTFS, because TTFS spikes encode non-binary values through spike timing.

To remove this bottleneck, we quantize the key ($K$) and value ($V$) projections to one bit, i.e., $\{+1,-1\}$. The dot product between a TTFS-encoded query ($Q$) and a binary $K$ or $V$ vector can then be implemented using only selective additions and subtractions. This avoids explicit multiplications while preserving the sparsity benefit of TTFS.

    \begin{figure}[htb]
      \centering
        \includegraphics[width=0.49\textwidth]{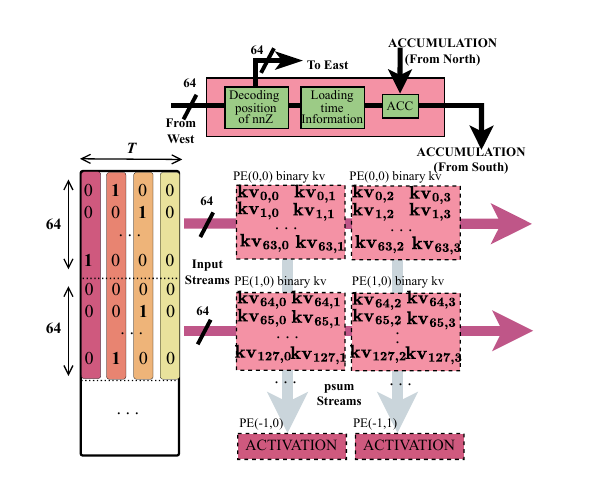}
      \caption{Architecture design for scores calculations.}
      \label{fig:scores_cal}
  \end{figure}

To support this 1-bit attention efficiently, we adopt a PE-array dataflow inspired by Canon~\cite{bai2025data}. During inference, the binary $K$/$V$ vectors are preloaded into local PE memory, and the TTFS-encoded query stream is broadcast to the PE array. Each PE accumulates its local binary values only at the time steps indicated by incoming spikes, and partial sums are propagated across PEs for final reduction. This design minimizes data movement and exploits the spatio-temporal sparsity of TTFS inputs, as shown in Figure~\ref{fig:scores_cal}.

    \begin{figure}[htb]
      \centering
        \includegraphics[width=0.49\textwidth]{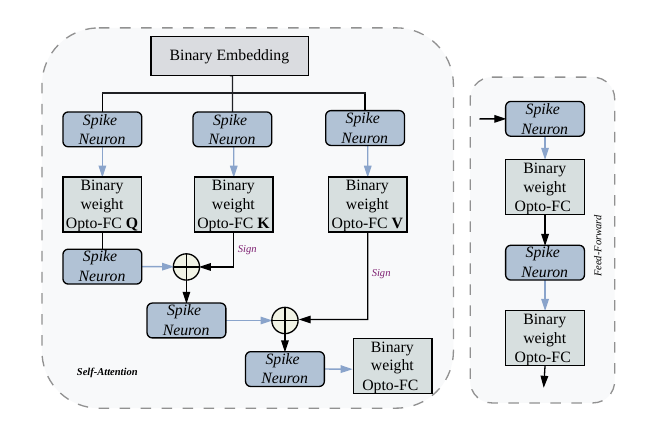}
        \caption{Otters++-based transformer structure.}
      \label{fig:main_structure}
  \end{figure}

Together with feed-forward layers built from Opto-FC and the spiking neuron in Section~\ref{subsec:Otters++_neuron}, this forms the complete Otters++ Transformer architecture, as shown in Figure~\ref{fig:main_structure}.

\subsection{QNN-Equivalent Construction of Otters++}
\label{subsec:qnn_equiv}

To make device-faithful training possible, we construct each Otters++ layer as the physical counterpart of an $n$-bit QNN layer. Consider an unsigned QNN layer with quantized input
\begin{equation}
q_i^{\,l-1} =
\operatorname{Clip}\!\left(
\left\lfloor \frac{a_i^{\,l-1}}{\alpha^{\,l-1}} \right\rfloor,
0, T
\right),
\qquad
x_{q,i}^{\,l-1} = \alpha^{\,l-1} q_i^{\,l-1},
\label{eq:qnn_input}
\end{equation}
where $\alpha^{\,l-1}>0$ is the learnable step size and
\begin{equation}
T = 2^n - 1.
\label{eq:T_definition}
\end{equation}

We define the QNN-to-Otters++ construction through four conditions:

\begin{enumerate}
    \item The SNN time window uses $T=2^n-1$ discrete levels, matching the positive quantization range of the $n$-bit QNN.
    \item Each logical level $k \in \{0,1,\dots,T-1\}$ is mapped to a physical spike time $t_k$ such that
    \begin{equation}
    O(t_k) = \frac{T-k}{T}.
    \label{eq:time_mapping}
    \end{equation}
    \item The physical synaptic scaling factor is set as
    \begin{equation}
    \gamma_{ij}^{l} = w_{ij}^{l}\, \alpha^{\,l-1} T.
    \label{eq:gamma_mapping}
    \end{equation}
    \item The firing threshold is defined as a step-wise decreasing function:
    \begin{equation}
    \theta^{l}(t) = \alpha^{l}(T-k),
    \qquad
    t_k \le t < t_{k+1}.
    \label{eq:threshold_mapping}
    \end{equation}
\end{enumerate}

\noindent\textbf{Proposition 1.}
\emph{Under the construction in Eqs.~(\ref{eq:time_mapping})--(\ref{eq:threshold_mapping}), an Otters++ layer is functionally equivalent to the corresponding QNN layer. Specifically, the membrane potential of each SNN neuron equals the QNN pre-activation, and the integer encoded by the SNN output spike time equals the clipped quantized output of the QNN.}

%The proposition follows from two simple observations. First, Eq.~(\ref{eq:time_mapping}) ensures that the device response at time $t_k$ equals the normalized quantization level. Combined with Eq.~(\ref{eq:gamma_mapping}), each presynaptic spike contributes exactly the weighted QNN input term. Therefore, the accumulated membrane potential equals the QNN pre-activation. Second, Eq.~(\ref{eq:threshold_mapping}) makes the earliest firing time implement the same floor-and-clip rule as unsigned quantization. As a result, the Otters++ forward pass becomes a device-faithful realization of the corresponding QNN layer.

\noindent\textbf{Proof of Proposition 1.} We prove equivalence in two phases: (I)~the integration phase, showing that the SNN membrane potential equals the QNN
  pre-activation; and (II)~the firing phase, showing that the SNN spike-time encoding recovers the QNN's quantized integer output.
  Together they establish a bijection between the two representations.

  \subsubsection{Phase I Integration Equivalence}
  \label{sec:proof_integration}

  Consider an unsigned $n$-bit QNN layer $l$. Its quantized input from the previous layer is:
  \begin{equation}
      x_{q,i}^{l-1} = \alpha^{l-1} \cdot q_i^{l-1}, \qquad
      q_i^{l-1} = \mathrm{Clip}\!\left(\left\lfloor \frac{a_i^{l-1}}{\alpha^{l-1}} \right\rfloor, 0, T\right)
  \end{equation}
  where $\alpha^{l-1} > 0$ is the learnable step size and $q_i^{l-1} \in \{0, 1, \dots, T\}$ is the integer quantization index.

  \paragraph{Towards TTFS encoding of the integer index}
  We encode $q_i^{l-1}$ as a spike time via the mapping:
  \begin{equation}
  \label{eq:ttfs_encoding}
      k_i = T - q_i^{l-1}
  \end{equation}
  This is a valid TTFS encoding: a larger quantized value $q_i^{l-1}$ produces a smaller (earlier) time index $k_i$, consistent with
   the ``earlier spike $\Rightarrow$ stronger activation'' principle. The boundary cases are: $q_i^{l-1} = T$ maps to $k_i = 0$
  (earliest spike); $q_i^{l-1} = 0$ maps to $k_i = T$ (no spike within the active window).

  \paragraph{ Device response recovers the quantization index}
  By Condition~2 of Proposition~1, the device output at time $t_{k_i}$ is:
  \begin{equation}
      O(t_{k_i}) = \frac{T - k_i}{T} \overset{\eqref{eq:ttfs_encoding}}{=} \frac{q_i^{l-1}}{T}
  \end{equation}
  Hence the memristive decay function linearly encodes the integer quantization index into a normalized output in $[0, 1]$.

  \paragraph{PSP equals weighted QNN input}
  The post-synaptic potential contribution from synapse $(i \to j)$ is defined as:
  \begin{align}
      \epsilon'(w_{ij}^l,\, t_{k_i})
      &= \gamma_{ij}^l \cdot O(t_{k_i}) \nonumber \\
      &= \bigl(w_{ij}^l \cdot \alpha^{l-1} \cdot T\bigr) \cdot \frac{q_i^{l-1}}{T} \nonumber \\
      &= w_{ij}^l \cdot \underbrace{\alpha^{l-1} \cdot q_i^{l-1}}_{= \, x_{q,i}^{l-1}}
      \label{eq:psp_equiv}
  \end{align}
  where we used Condition~3 in the second line. Thus, a single SNN spike produces exactly the same contribution as the corresponding
   weighted input in the QNN.

  \paragraph{Membrane potential equals pre-activation}
  Summing over all presynaptic neurons and adding the bias:
  \begin{equation}
  \label{eq:membrane_equiv}
      V_j^l = \sum_{i} \epsilon'(w_{ij}^l,\, t_{\mathrm{spike},i}^{l-1}) + b_j^l
      = \sum_{i} w_{ij}^l \, x_{q,i}^{l-1} + b_j^l
      = a_j^l
  \end{equation}
  This establishes that the SNN membrane potential is identical to the QNN pre-activation. \hfill$\square$

  \subsubsection{Phase II: Firing Equivalence}
  \label{sec:proof_firing}

  We now show that the integer value encoded by the SNN output spike time equals the QNN's clipped floor output:
  \begin{equation}
  \label{eq:firing_goal}
      q_{\mathrm{out},j}^l \triangleq T - k_{\mathrm{fire}} = \mathrm{Clip}\!\left(\left\lfloor \frac{a_j^l}{\alpha^l}
  \right\rfloor, 0, T\right) = q_j^l
  \end{equation}

  \paragraph{Firing condition}
  The neuron fires at the earliest time step $k$ satisfying $V_j^l \geq \theta^l(t_k)$. Substituting $V_j^l = a_j^l$ from
  \eqref{eq:membrane_equiv} and Condition~4:
  \begin{equation}
  \label{eq:fire_condition}
      a_j^l \geq \alpha^l \cdot (T - k)
  \end{equation}

  \paragraph{Earliest firing time}
  Since $\alpha^l > 0$, the threshold $\alpha^l(T-k)$ is monotonically decreasing in $k$. Therefore, if the condition is met at some
   $k^*$, it is also met for all $k > k^*$. The TTFS rule selects the earliest (smallest) such $k$, equivalently the largest integer
   $(T-k)$ satisfying $T-k \leq a_j^l / \alpha^l$. By definition of the floor function:
  \begin{equation}
  \label{eq:floor_equiv}
      T - k_{\mathrm{fire}} = \left\lfloor \frac{a_j^l}{\alpha^l} \right\rfloor
  \end{equation}
  provided the result lies in $\{0, 1, \ldots, T\}$.

  \paragraph{ Boundary conditions naturally implement clipping}
  \begin{itemize}
      \item \textbf{Upper saturation} ($\lfloor a_j^l / \alpha^l \rfloor > T$): The condition \eqref{eq:fire_condition} holds for
  all $k \in \{0,\ldots,T{-}1\}$. The neuron fires at $k_{\mathrm{fire}} = 0$, encoding the value $T - 0 = T$. This implements
  $\min(\cdot, T)$.

      \item \textbf{Lower saturation} ($\lfloor a_j^l / \alpha^l \rfloor < 0$): The ratio $a_j^l / \alpha^l < 0$ while $T-k \geq 1$
  for all $k < T$, so \eqref{eq:fire_condition} is never satisfied. The neuron does not fire, encoding the value $0$. This
  implements $\max(\cdot, 0)$.
  \end{itemize}

  Combining all cases:
  \begin{equation}
  \label{eq:full_equiv}
      q_{\mathrm{out},j}^l = \mathrm{Clip}\!\left(\left\lfloor \frac{a_j^l}{\alpha^l} \right\rfloor, 0, T\right) = q_j^l
  \end{equation}
  This completes the proof that the SNN firing mechanism produces the same quantized integer output as the QNN.   \hfill$\square$    
\subsection{Hybrid SNN-Forward/QNN-Backward Training}
\label{subsec:hybrid_training}

Proposition~1 makes the QNN more than a pretraining surrogate: it becomes the differentiable counterpart of Otters++ itself. This allows us to use a hybrid \emph{SNN-forward/QNN-backward} training strategy.

In the forward pass, we preserve the true Otters++ computation flow. Activations are first converted into spike times through threshold comparison, then mapped to discretized physical times $\{t_k\}$, passed through the device response, and finally reconstructed into layer outputs. This keeps training consistent with the actual hardware execution path.

In the backward pass, instead of differentiating through sparse and discontinuous first-spike events, we compute gradients through the equivalent QNN path using the straight-through estimator (STE). For the activation path, we use
\begin{equation}
\frac{\partial \mathcal{L}}{\partial \tilde{x}^{\,l}}
=
\mathbf{1}_{\,0 \le \tilde{x}^{\,l}/\hat{\alpha}^{\,l} \le Q_p}
\frac{\partial \mathcal{L}}{\partial z^{\,l}},
\label{eq:ste_x}
\end{equation}
and for the step size,
\begin{equation}
\frac{\partial \mathcal{L}}{\partial \alpha^{\,l}}
=
\frac{1}{\sqrt{n \cdot Q_p}}
\sum
\left[
\mathbf{1}_{\text{over}} \cdot Q_p
+
\mathbf{1}_{\text{mid}} \cdot
\left(\lfloor q \rfloor - q\right)
\right]
\frac{\partial \mathcal{L}}{\partial z^{\,l}},
\label{eq:ste_alpha}
\end{equation}
where
\begin{equation}
q = \frac{\tilde{x}^{\,l}}{\hat{\alpha}^{\,l}},
\qquad
\mathbf{1}_{\text{over}} = \mathbf{1}_{q > Q_p},
\qquad
\mathbf{1}_{\text{mid}} = \mathbf{1}_{0 \le q \le Q_p}.
\end{equation}

\noindent\textbf{Corollary 1.}
\emph{Under nominal device conditions, the layer-wise Otters++ forward pass and the corresponding QNN forward pass are identical. Therefore, STE gradients computed through the QNN path provide valid descent directions for the Otters++ parameters $\{W_l,\beta_l,\alpha_l\}$.}

To further stabilize optimization, we adopt knowledge distillation from a pretrained teacher. The student is supervised at both the output level and the hidden-state level:
\begin{equation}
\mathcal{L}
=
\mathrm{KL}\!\left(
\sigma(\hat{\mathbf{y}}^S/\tau)\,\|\,\sigma(\hat{\mathbf{y}}^T/\tau)
\right)
+
\sum_{l=0}^{L}
\left\|
\mathbf{h}_l^S - \mathbf{h}_l^T
\right\|_2^2,
\label{eq:kd_loss}
\end{equation}
where $\hat{\mathbf{y}}^T$ and $\hat{\mathbf{y}}^S$ are the teacher and student logits, $\tau$ is the temperature, and $\mathbf{h}_l^T$ and $\mathbf{h}_l^S$ are the hidden states at layer $l$. This distillation objective complements the hybrid training framework by improving optimization under quantization and spike-timing constraints. Algorithm~\ref{alg:ttfs-kd} summarizes the full training procedure.

\begin{algorithm}[t]
\caption{Noise-Robust Otters++ Training via Knowledge Distillation}
\label{alg:ttfs-kd}
\begin{algorithmic}[1]
\REQUIRE Pretrained teacher $f^T$; student $f^S$ initialized from $f^T$; training set $\mathcal{D}$; epochs $E$; learning rates $\eta$ and $\eta_\alpha$; temperature $\tau$; weight bits $b_w=1$; activation bits $b_a$; time resolution $\delta$
\ENSURE Noise-robust Otters++ model $f^S$

\STATE Initialize learnable step sizes $\{\alpha_l\}$ for each layer $l$
\STATE $T \leftarrow 2^{b_a}-1$, \quad $Q_n \leftarrow 0$, \quad $Q_p \leftarrow 2^{b_a}-1$

\FOR{epoch $=1$ to $E$}
    \FOR{each mini-batch $(\mathbf{X},\mathbf{y}) \in \mathcal{D}$}
        \STATE $\hat{\mathbf{y}}^T,\{\mathbf{h}_l^T\}_{l=0}^{L} \leftarrow f^T(\mathbf{X})$ \textbf{with no gradient}
        \FOR{each layer $l$ in $f^S$}
            \STATE $\mathbf{W}_l^q \leftarrow \operatorname{sign}(\mathbf{W}_l-\bar{\mathbf{W}}_l)\cdot \frac{\|\mathbf{W}_l\|_1}{n_l}$ \hfill \textit{// BWN}
            % \STATE $\tilde{\mathbf{x}}_l \leftarrow \mathbf{x}_l + \boldsymbol{\beta}_l$ \hfill \textit{// learnable bias shift}
            % \STATE $\hat{\alpha}_l \leftarrow 2^{\lfloor \log_2 \alpha_l \rceil}$ \hfill \textit{// power-of-2 step size}
            \FOR{$k=0$ to $T-1$}
                \STATE $\theta_k \leftarrow \alpha_l (T-k)$ \hfill \textit{// unsigned threshold}
                \STATE $t_k \leftarrow \left\lfloor O^{-1}\!\left(\frac{T-k}{T}\right)/\delta \right\rceil \cdot \delta$ \hfill \textit{// discretized spike time}
            \ENDFOR
            \STATE Assign $\hat{t}_{ij} \leftarrow t_k$ for the first $k$ such that $\tilde{x}_{ij} \ge \theta_k$ \hfill \textit{// priority encoding}
            \STATE $o_{ij}= \operatorname{Clamp}_{[\ell_{ij},h_{ij}]} \left( O_{\mathrm{mean}}(\hat{t}_{ij})+\epsilon_{ij} \right)$ \hfill \textit{// device noise injection}
            \STATE $\mathbf{z}_l \leftarrow \mathbf{W}_l^q \big(\alpha_l \cdot T \cdot \mathbf{o}\big)$ \hfill \textit{// layer output}
        \ENDFOR
        \STATE Compute $\mathcal{L}$ using Eq.~(\ref{eq:kd_loss})
        \STATE Compute $\nabla \mathcal{L}$ via STE in Eqs.~(\ref{eq:ste_x})--(\ref{eq:ste_alpha})
        \STATE Update $\{\mathbf{W}_l,\boldsymbol{\beta}_l\} \leftarrow \text{BertAdam}(\eta)$; \quad $\{\alpha_l\} \leftarrow \text{BertAdam}(\eta_\alpha)$
    \ENDFOR
\ENDFOR

\RETURN $f^S$
\end{algorithmic}
\end{algorithm}

\subsection{Noise-Aware Extension under Measured Device Variation}
\label{subsec:noise_aware}

The equivalence in Proposition~1 holds for the nominal device response curve. In practice, however, the measured optoelectronic response exhibits run-to-run variation. To make training robust to this effect, we extend the forward pass to sample from the measured variation envelope.

    \begin{figure}[htb]
      \centering
        \includegraphics[width=0.49\textwidth]{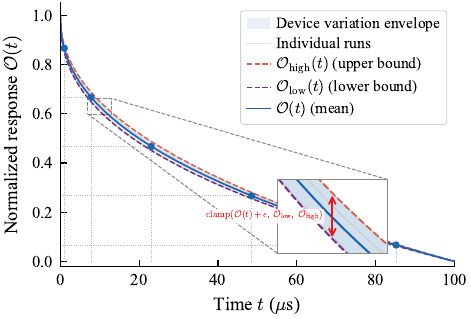}
      \caption{Noise decay function under measurement.}
      \label{fig:noise_decay}
  \end{figure}

Specifically, for a discretized spike time $\hat{t}_{ij}$, we sample the device response as
\begin{equation}
o_{ij}
=
\operatorname{Clamp}_{[\ell_{ij},h_{ij}]}
\left(
O_{\mathrm{mean}}(\hat{t}_{ij})+\epsilon_{ij}
\right),
\qquad
\epsilon_{ij}\sim\mathcal{N}(0,\sigma_{ij}^2)
\label{eq:noise_sampling}
\end{equation}

where $\ell_{ij}=O_{\mathrm{low}}(\cdot)$ and $h_{ij}=O_{\mathrm{high}}(\cdot)$ are the lower and upper measured response bounds, respectively. This perturbs the nominal SNN forward pass while keeping the perturbation physically grounded.
Under this model, the perturbed membrane potential becomes
\begin{equation}
\tilde{V}_j^{\,l}
=
a_j^{\,l}
+
\sum_i
w_{ij}^{\,l}\, \alpha^{\,l-1} T \,\Delta o_{ij},
\qquad
\Delta o_{ij} = o_{ij} - O(\hat{t}_{ij}).
\label{eq:noisy_membrane}
\end{equation}
The backward pass still uses the nominal QNN graph and the STE in Section~\ref{subsec:hybrid_training}. This makes the method efficient while minimizing the expected loss under realistic device uncertainty. In effect, the model is trained not only to match the nominal device response, but also to remain robust within the measured variation envelope.

\subsection{Otters++ Energy Calculation}
\label{subsec:otters_energy_calculation}

We next derive the energy model of Otters++ under the same accounting framework. Unlike conventional multi-step SNNs, Otters++ does not digitally evaluate the TTFS decay term at every synaptic event. Instead, the temporal component is directly realized by the measured analog response of the optoelectronic device. Therefore, its energy is dominated by three parts: 1) thresholding for first-spike generation, 2) spike-driven analog accumulation, and 3) binary K/V write or read overhead in the attention pipeline.

For consistency with the previous subsections, we decompose the total energy into the projection energy $E_{\text{fc}}$ and the attention-score energy $E_{\text{score}}$. The projection-layer energy is given by
\begin{align}
E_{\text{fc}} = & B \cdot S \cdot C_o \cdot \Bigl[ \underbrace{T \cdot \bigl( E_{\text{CMP}} + E^{\text{Read}}_{\text{threshold}} \bigr)}_{\text{Thresholding}} + \underbrace{E^{\text{Write}}_{\text{binarykv}}}_{\text{K/V Write}} \nonumber \\
& + \underbrace{C_i \cdot T \cdot \Bigl( s_r \cdot \bigl( E_{\text{ACC}} + E^{\text{Read}}_{\text{Analog}} + E^{\text{sparse}}_{\text{move}} \bigr) + E_{\text{leakage}} \Bigr)}_{\text{Spike Processing}} \Bigr] \\[15pt]
E_{\text{score}} = & B \cdot h \cdot S^2 \cdot \Bigl[ \underbrace{T \cdot \bigl( E_{\text{CMP}} + E^{\text{Read}}_{\text{threshold}} \bigr)}_{\text{Thresholding}} +d_k \cdot T\cdot \nonumber \\
&  \underbrace{ \Bigl( s_r \cdot \bigl( E_{\text{ACC}} + E^{\text{Read}}_{\text{Analog}} + E^{\text{sparse}}_{\text{move}} + E^{\text{Read}}_{\text{binarykv}} \bigr) + E_{\text{leakage}} \Bigr)}_{\text{Spike Processing}} \Bigr]
\end{align}

These equations can be interpreted as follows. The prefactors $B \cdot S \cdot C_o$ and $B \cdot h \cdot S^2$ count the total number of output neurons in the projection layer and the total number of pairwise attention-score evaluations across all heads, respectively. For each output, the first term models \emph{thresholding}, i.e., the step-wise comparison between the membrane state and the time-varying threshold over $T$ time steps, together with the threshold read cost. The second term accounts for the write-back of binarized key/value states in the projection path. The third term captures \emph{spike processing}: for each presynaptic channel and each time step, an active event incurs an accumulation, an analog read from the device response, and sparse spike movement, scaled by the average spike rate $s_r$, while the leakage term accounts for the state-holding cost during temporal evolution.

Compared with conventional SNNs, the key difference is that Otters++ replaces explicit digital temporal decoding and weight-modulated MAC-style evaluation by a device-native analog response. As a result, the dominant synaptic cost is reduced to event-driven accumulation plus analog readout, without separately paying for digital evaluation of the TTFS temporal function. Moreover, unlike full-precision and quantized Transformers, Otters++ communicates binary spike, which further reduces data-movement overhead. Therefore, the energy advantage of Otters++ comes from jointly reducing arithmetic complexity, activation precision, and temporal decoding cost.

\section{Results}
\label{sec:results}

In this section, we evaluate Otters++ on seven datasets from the GLUE benchmark. We compare its performance against both standard QNN and SNN baselines, using $\text{BERT}_{\text{base}}$ as the teacher model for knowledge distillation. We further provide a detailed analysis of the model's energy efficiency and robustness. All experiments were conducted on three NVIDIA A100 GPUs with a fixed 4-bit simulation window size recommended by Sorbet~\cite{tangsorbet} (In our setting, it is equal to timestep $T=15$). 

\subsection{Fabricated In$_2$O$_3$ TFTs }
\label{sec:Fabricated }
To prepare the indium oxide thin-film transistors (In$_2$O$_3$ TFTs), indium nitrate was first dissolved in a mixed solvent of 2-methoxyethanol (2-ME), acetylacetone (AcAc), and ammonium hydroxide (NH$_3$·H$_2$O) to form a precursor solution, which was stirred overnight to ensure complete dissolution and coordination. Subsequently, gate electrodes were fabricated on a silicon substrate coated with a SiO$_2$ insulating layer, followed by sequential deposition of 8~nm chromium and 50~nm gold via electron-beam evaporation. A 30~nm-thick Al$_2$O$_3$ dielectric layer was then uniformly deposited over the substrate using atomic layer deposition (ALD). The In$_2$O$_3$ precursor solution was spin-coated onto the dielectric surface, after which the channel regions were defined through standard photolithography, and the unprotected areas were removed by hydrochloric acid wet etching. The films were annealed in air at 300~$^\circ$C for 1~hour to enhance crystallinity and improve film quality. Portions of the Al$_2$O$_3$ layer were subsequently etched to expose selected regions of the gate electrodes. Finally, source and drain electrodes, along with interconnects, were patterned and metallized with an additional 8~nm chromium and 50~nm gold layer via electron-beam evaporation. The indium oxide thin-film transistor was characterized under a gate bias of 0~V and a drain bias of 5~mV. Upon 405~nm laser illumination, oxygen vacancies in the channel layer were photoionized, generating free electrons and thereby enhancing the channel conductivity. 
% \begin{figure}[h!]
%     \centering
%     \begin{subfigure}[b]{0.49\textwidth}
%         \centering
%     \includegraphics[width=0.68\textwidth]{devic.png}
%         \caption{Optical microscope}
%     \end{subfigure}
%     \hfill
%     \begin{subfigure}[b]{0.49\textwidth}
%         \centering
%         \includegraphics[width=1.09\textwidth]{figures/4-4.pdf}
%         \caption{Otters under optical microscope}
%         \label{fig:optical}
%     \end{subfigure}
%     \caption{Details design of Otters}
% \end{figure}

\begin{figure}[htbp]
    \centering
    \includegraphics[width=1\linewidth]{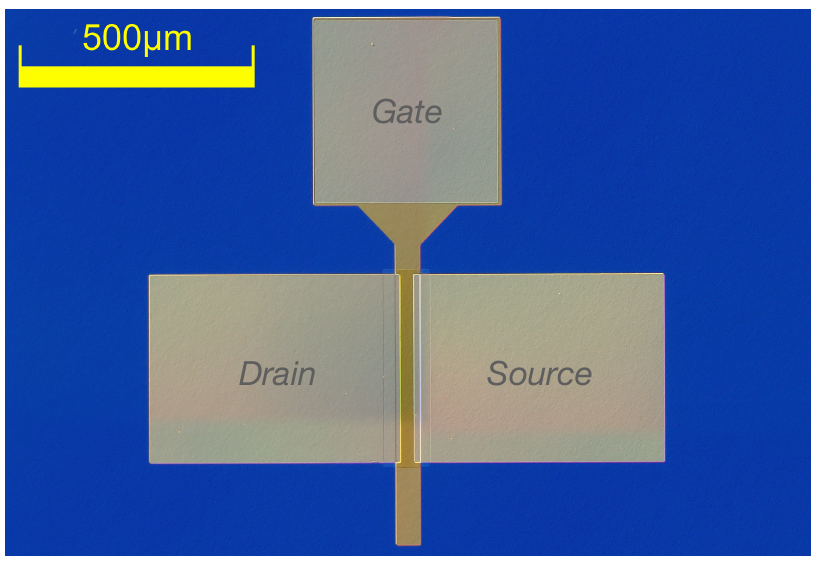}
    \caption{Details design of Otters++.}
    \label{fig:optical}
\end{figure}

Regarding integration with state-of-the-art digital hardware, recent studies show that In$_2$O$_3$ TFTs are compatible with 3D stacking on CMOS chips. This is because they have a low thermal budget ($\leq$ 300 °C) and good uniformity~\cite{tang2022monolithic}. This provides a practical path for vertical integration. However, several challenges remain, such as managing wire density in stacked layers, aligning the photodetectors/TFT layers with metal layers, and handling device variability in large arrays.

\subsection{GLUE Benchmark Performance}

Table~\ref{tbl:glue_performance_avg} summarizes the GLUE benchmark results. We report two variants of Otters++. \emph{Nominal} uses the mean measured device response. \emph{Measured Variation} samples run-to-run device responses from the experimentally measured variation envelope, and the reported results are mean$\pm$std over 20 independent trials.

Under the nominal setting, Otters++ achieves the best average score among all SNN-based methods, reaching 84.17. It outperforms SpikingBERT, SpikeLM, 1-bit SpikeLM, and 1-bit Sorbet by 3.34, 3.66, 3.93, and 4.37 points, respectively. Compared with the original Otters, Otters++ improves the average score from 83.22 to 84.17, with gains on six out of seven tasks: QQP, MNLI-m, SST-2, QNLI, MRPC, and STS-B, while matching the same RTE score. The largest improvements are observed on QNLI (+1.46) and MRPC (+2.44), suggesting that the proposed hybrid training framework better aligns the physical TTFS forward path with the optimization objective.

Under measured device variation, Otters++ maintains the same average score of 84.17, with a standard deviation of 0.28 across 20 independent variation trials. The task-level results show that the model remains stable under realistic hardware uncertainty: performance improves over the nominal setting on QQP, MNLI-m, QNLI, and RTE, while showing moderate decreases on SST-2, MRPC, and STS-B. Importantly, the average performance is not degraded by measured variation, indicating that the proposed training procedure preserves robustness under experimentally observed device fluctuations.

%Under measured device variation, Otters++ maintains the same average score of 84.17, with a standard deviation of 0.28 across variation trials. The task-level results show that the model remains stable under realistic hardware uncertainty: performance improves over the nominal setting on QQP, MNLI-m, QNLI, and RTE, while showing moderate decreases on SST-2, MRPC, and STS-B. Importantly, the average performance is not degraded by the measured variation, indicating that the variation-aware training strategy improves robustness without sacrificing overall accuracy.

\begin{table*}[hbt]
\centering
\caption{Performance comparison on the GLUE benchmark. All scores are accuracy, except for STS-B (Pearson correlation). “*” indicates that the model size was not reported in the original paper. \textbf{Bold} indicates the best performance among SNN models. Only Otters++ quantizes KV to 1 bit. ``Nominal'' denotes evaluation using the mean device response, while ``Variation'' denotes evaluation under measured run-to-run device variation sampled from the experimental response envelope.}
\label{tbl:glue_performance_avg}
\resizebox{0.9\textwidth}{!}{%
\begin{tabular}{@{}lcccccccccc@{}}
\toprule
\textbf{Model} & \textbf{Size} & \textbf{QQP} & \textbf{MNLI-m} & \textbf{SST-2} & \textbf{QNLI} & \textbf{RTE} & \textbf{MRPC} & \textbf{STS-B} & \textbf{Average} \\
\midrule
$\text{BERT}_{\text{base}}$~\cite{devlin2019bert} & 418M & 91.3 & 84.7 & 93.3 & 91.7 & 72.6 & 88.2 & 89.4 & 87.31 \\
DistilBERT~\cite{sanh2020distilbertdistilledversionbert} & 207M & 88.5 & 82.2 & 91.3 & 89.2 & 59.9 & 87.5 & 86.9 & 83.64 \\
$\text{TinyBERT}_6$~\cite{jiao2020tinybert} & 207M & - & 84.6 & 93.1 & 90.4 & 70.0 & 87.3 & 83.7 & 84.85 \\
Q2BERT~\cite{zhang2020ternarybert}      & 43.0M & 67.0 & 47.2 & 80.6 & 61.3 & 52.7 & 68.4 & 4.4 & 54.51 \\
BiT~\cite{liu2022bit}     & 13.4M & 82.9 & 77.1 & 87.7 & 85.7 & 58.8 & 79.7 & 71.1 & 77.57 \\
SpikingFormer~\cite{zhou2023spikingformer}  & * & 83.8 & 67.8 & 82.7 & 74.6 & 58.8 & 74.0 & 72.3 & 73.43 \\
SpikingBERT~\cite{bal2024spikingbert}  & 50M & 86.8 & 78.1 & 88.2 & 85.2 & 66.1 & 79.2 & 82.2 & 80.83 \\
SpikeLM~\cite{xingspikelm}   & * & 87.9 & 76.0 & 86.5 & 84.9 & 65.3 & 78.7 & 84.3 & 80.51 \\
\midrule
1-bit SpikeLM ~\cite{xingspikelm} & * & 87.2 & 74.9 & 86.6 & 84.5 & 65.7 & 78.9 & 83.9 & 80.24 \\
1-bit Sorbet~\cite{tangsorbet} & 13.4M & 86.5 & 77.3 & 90.4 & 86.1 & 60.3 & 79.9 & 78.1 & 79.80 \\

\text{Otters~\cite{yan2026otters}} & 13.4M & \text{87.67} & \text{78.50} & \text{91.28} & \text{86.42} & \text{68.95} & \text{84.56} & \text{85.19} & \text{83.22} \\
\textbf{Otters++ (Nominal)} & 13.4M & 88.03 & 79.56 &91.97
      & 87.88 & 68.95 & 87.00
      & 85.79 & 84.17 \\
            \textbf{Otters++ (Variation)}  & 13.4M                                                                                                        
        & 88.24$\pm$0.07 & 79.93$\pm$0.23 & 90.53$\pm$0.57                                                                
        & 88.19$\pm$0.18 & 71.28$\pm$1.54 & 86.09$\pm$0.99                                                                
        & 84.93$\pm$0.38 & 84.17$\pm$0.28 \\                
\bottomrule
\end{tabular}}
\end{table*}

\subsection{Energy Efficiency Analysis}
\label{subsec:energy_efficiency}

We next evaluate the system-level energy efficiency of Otters++ from three perspectives: overall per-layer energy, component-wise energy breakdown, and sensitivity to communication hops. 
Energy costs are derived from a combination of 22-nm post-synthesis measurements and prior hardware studies. For FP32 operations, we use a MAC energy of 4.6 pJ and an FP32 arithmetic energy of 0.9 pJ from prior literature~\cite{horowitz20141}, since our synthesis flow targets integer datapaths. For INT4 models, we distinguish between 4-4-16-bit MACs (0.0848 pJ) and 1-4-16-bit MACs (0.0663 pJ). The energies of 4-16-16-bit, 2-16-16-bit, and 1-16-16-bit ACC operations are 0.0502 pJ, 0.0477 pJ, and 0.0429 pJ, respectively. SNN-specific comparison and subtraction operations are each modeled as 0.0502 pJ. The analog read energy is 0.0246 pJ, including 0.00875 pJ for powering the TFT, 1.33$\times 10^{-6}$ pJ for sampling, 0.0053 pJ for the ADC and amplifier, and 0.0105 pJ for the 4-bit LUT. We further model static leakage as 0.002 pJ per cycle, weight/activation read-write energy as 0.0985 pJ/bit, and sparse data movement as 0.18 pJ/bit. Integer compute, data movement, and memory-access costs are obtained from a commercial 22-nm process, while FP32 compute and ADC costs are taken from existing literature~\cite{horowitz20141,su20235}.

%The energy model is configured for a BERT-base architecture with a batch size ($B$) of 64, a sequence length ($S$) of 128, and input/output channel dimensions ($C_i, C_o$) of 768. The model features 12 attention heads ($h$), with a per-head dimension ($d_k$) of 64. Energy costs are derived from established models. For FP32 operations, we assume that a multiply-accumulate (MAC) consumes 4.6~pJ and a clamp operation consumes 0.9~pJ which is from ~\cite{horowitz20141} cause our platform doesn't support FP calculations. For our INT4 models, we differentiate between a 4-4-16bits MAC (0.0848~pJ) and a 1-4-16bits MAC (0.0663~pJ). The costs for 4-16-16bits, 2-16-16bits and 1-16-16bits ACC are 0.0502~pJ, 0.0477~pJ and 0.0429~pJ, respectively. 
%SNN-specific operations, such as comparison and subtraction, are each modeled at 0.0502~pJ. The total energy for an analog read operation is 0.0246~pJ (0.00875pJ for power the TFT, 1.33e-6pJ for sampling, 0.0053pJ for ADC including the amplifier and 0.010505 for the 4bits LUT). 
%The model accounts for a static leakage energy \(E_{\text{leakage}}\) of 0.002~pJ per cycle, a weight activation (read/write) cost of 0.0985~pJ/bit, and a sparse data movement cost of 0.18~pJ per bit. 
%All integer compute, data movement, and memory access costs are based on measurements from a commercial 22nm process, while the floating-point compute energy~\cite{horowitz20141} and ADC values~\cite{su20235} are taken from existing literature.

\subsubsection{Overall Energy Comparison}
\begin{figure}[htbp]
    \centering
    \includegraphics[width=1\linewidth]{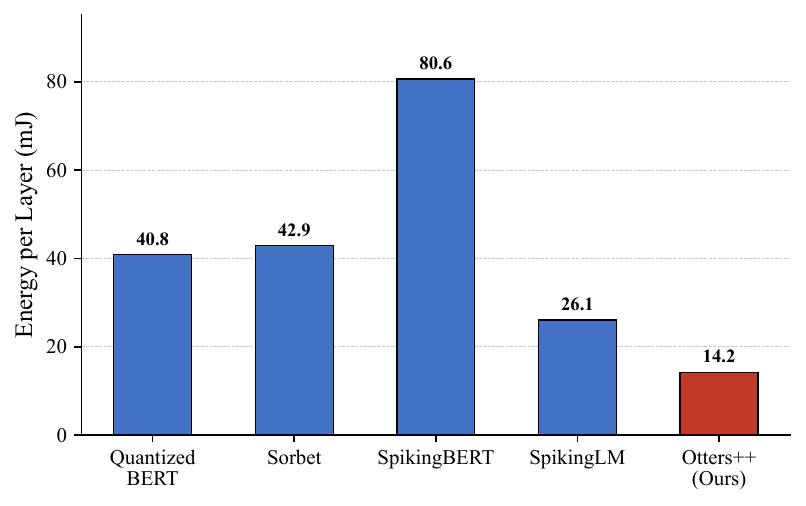}
    \caption{Per-layer energy comparison under the unified hardware cost model. Otters++ achieves the lowest per-layer energy among all compared methods.}
    \label{fig:energy_per_layer}
\end{figure}

Figure~\ref{fig:energy_per_layer} compares the per-layer energy of different low-bit language models under the same hardware cost model. Otters++ achieves the lowest energy consumption, requiring only 14.2 mJ per layer, compared with 40.8 mJ for Quantized BERT, 42.9 mJ for Sorbet, 80.6 mJ for SpikingBERT, and 26.1 mJ for SpikingLM. This corresponds to energy reductions of 65.2\%, 66.9\%, 82.4\%, and 45.6\%, respectively. Equivalently, Otters++ is about 2.87$\times$, 3.02$\times$, 5.68$\times$, and 1.84$\times$ more energy-efficient than these baselines. These results show that the proposed design substantially lowers the inference cost not only relative to conventional quantized Transformers, but also relative to prior SNN-based language models.

\subsubsection{Energy Breakdown Analysis}
\begin{figure}[htbp]
    \centering
    \includegraphics[width=1\linewidth]{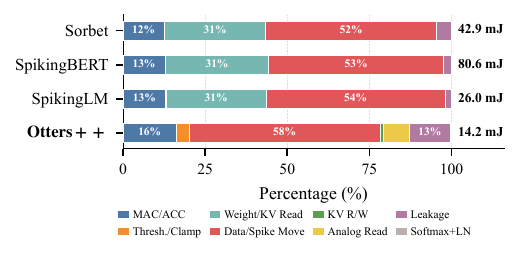}
    \caption{Per-layer energy breakdown across models. Otters++ removes the large weight/KV-read component that dominates prior baselines, while keeping analog-read overhead modest.}
    \label{fig:energy_breakdown}
\end{figure}

To better understand where the savings come from, Figure~\ref{fig:energy_breakdown} reports the normalized energy breakdown. For Sorbet, SpikingBERT, and SpikingLM, the dominant cost comes from data/spike movement, which accounts for 52\%--54\% of the total energy, followed by weight/KV read at 31\%, and MAC/ACC at 12\%--13\%. 
In contrast, Otters++ exhibits a markedly different profile. Although data/spike movement remains the largest term at 58\%, the high-precision digital weight/KV-read bottleneck is substantially reduced and partly replaced by low-cost analog readout. The added analog read contributes only a modest single-digit share, while the remaining energy is mainly distributed across MAC/ACC (16\%) and leakage (13\%).
%In contrast, Otters++ exhibits a markedly different profile. Although data/spike movement remains the largest term at 58\%, the large weight/KV read component is effectively removed, while the added analog read contributes only a modest single-digit share. The remaining energy is mainly distributed across MAC/ACC (16\%) and leakage (13\%). 
%This result is important for two reasons. First, it confirms that the benefit of Otters++ does not come from a single overly optimistic assumption; rather, it comes from reshaping the entire energy profile. Second, it shows that the cost of accessing the analog device response is much smaller than the digital weight/KV traffic that dominates prior designs. In other words, shifting the TTFS temporal modulation from digital evaluation to device physics reduces one of the largest bottlenecks in low-bit Transformer inference. 

%The Otters++ energy band also captures different analog-read assumptions induced by device sharing: stronger sharing amortizes the device access cost across more synaptic operations, while weaker sharing gives the upper-bound analog-read estimate.

\subsubsection{Impact of Communication Hops}
\begin{figure}[htbp]
    \centering
    \includegraphics[width=1\linewidth]{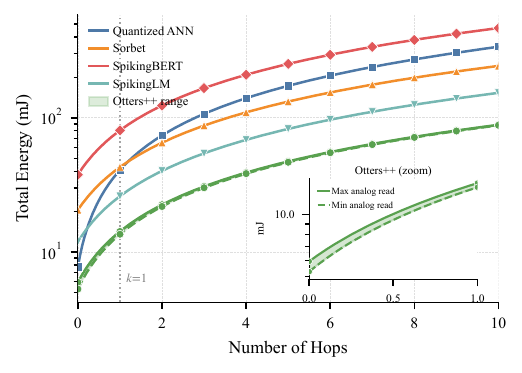}
    \caption{Total energy versus number of communication hops. The shaded region denotes the Otters++ energy range under minimum and maximum analog-read assumptions; the inset zooms in on the low-hop regime.}
    \label{fig:energy_vs_hops}
\end{figure}

Figure~\ref{fig:energy_vs_hops} further studies total energy as a function of the number of communication hops, which provides a more deployment-aware view than per-operation accounting alone. As expected, the energy of all methods increases monotonically with the hop count. However, the relative ordering remains unchanged: Otters++ consistently stays below all baselines from 0 to 10 hops. The shaded Otters++ band, bounded by the minimum- and maximum-analog-read assumptions, remains narrow, and the inset shows that this gap is still small even in the low-hop regime. This indicates that the advantage of Otters++ is robust to uncertainty in analog-read cost. More importantly, it shows that the proposed method retains its energy benefit even after communication overhead is explicitly incorporated into the model. Therefore, the efficiency gain of Otters++ is not limited to a favorable local-compute setting, but persists under increasingly communication-dominated deployment scenarios. The Otters++ energy band also captures different analog-read assumptions induced by device sharing: stronger sharing amortizes the device access cost across more synaptic operations, while weaker sharing gives the upper-bound analog-read estimate.

Overall, the three figures lead to a consistent conclusion. Otters++ achieves the lowest total energy, exhibits a more favorable energy composition by removing the large weight/KV-read bottleneck, and maintains this advantage as communication distance increases. These results support the main claim of this work: by implementing the TTFS temporal term through device physics, Otters++ reduces not only arithmetic cost, but also memory-access and system-level communication overhead.

\subsection{Ablation Study: Direct QNN-to-SNN Conversion vs. Integrated TTFS Training}
\label{subsec:ablation_qnn_vs_Otters++}

To isolate the effect of the proposed training strategy, we compare four settings in Table~\ref{tab:glue_results_qvss}: the full-precision BERT teacher, the quantized neural network (QNN), \textit{Otters} obtained by direct post-training QNN-to-SNN conversion, and \textit{Otters++}, which incorporates the TTFS forward path during training. This comparison examines whether the accuracy loss mainly comes from the physical TTFS representation itself, or from the mismatch between training in the QNN domain and deployment in the TTFS-SNN domain. Direct conversion introduces a small but consistent accuracy gap. The QNN achieves an average score of 83.81, while Otters drops to 83.22, corresponding to a 0.59-point degradation. The largest losses appear on RTE, MRPC, and STS-B, suggesting that direct conversion does not fully capture the discretized TTFS sampling process and the device-dependent response used during deployment. Thus, even when the QNN and SNN are functionally aligned under nominal conditions, training only in the QNN domain can still introduce a deployment-time mismatch.

Otters++ closes this gap by integrating the TTFS forward path into training. It improves the average score to 84.17, which is 0.95 points higher than Otters and 0.36 points higher than the QNN baseline. The gains over Otters are broad across tasks, including QQP, MNLI-m, SST-2, QNLI, MRPC, and STS-B, while maintaining the same RTE score. These results show that the proposed hybrid SNN-forward/QNN-backward training strategy not only reduces the QNN-to-SNN conversion gap, but also improves the final task performance under the nominal Otters++ execution path.

\begin{table*}[t]
  \centering
  \caption{Accuracy gap analysis on the GLUE benchmark under nominal device response. All scores are accuracy except for STS-B, which reports Pearson correlation.}
  \label{tab:glue_results_qvss}
  \resizebox{0.65\textwidth}{!}{%
  \begin{tabular}{lcccccccc}
    \toprule
    Model & QQP & MNLI-m & SST-2 & QNLI & RTE & MRPC & STS-B & Avg \\
    \midrule
    Full BERT
      & 91.30 & 84.70 & 93.30
      & 91.70 & 72.60 & 88.20
      & 89.40 & 87.31 \\
    QNN
      & 87.81 & 78.40 & 91.28
      & 86.80 & 70.03 & 86.51
      & 85.87 & 83.81 \\
    Otters
      & 87.67 & 78.50 & 91.28
      & 86.42 & 68.95 & 84.56
      & 85.19 & 83.22 \\
    \midrule
    Otters++
      & 88.03 & 79.56 & 91.97
      & 87.88 & 68.95 & 87.00
      & 85.79 & 84.17 \\
    \bottomrule
  \end{tabular}}
\end{table*}

\subsection{Ablation study of using Otters++ and traditional TTFS methods}

We further compare Otters++ with traditional TTFS implementations. Existing TTFS methods can be broadly divided into continuous-time and quantized-time designs. Continuous-time TTFS methods, such as TTFSFormer~\cite{zhaottfsformer}, propagate spike times between layers as FP32 values. This high-precision representation helps preserve accuracy after ANN-to-SNN conversion, but it also makes the hardware cost fundamentally different from low-bit TTFS. For example, an FP32 accumulator costs about 0.9 pJ per operation, whereas the INT4 accumulator used in our setting costs about 0.05 pJ. The wider representation also increases memory-access and data-movement energy.

To make the comparison more controlled, we re-evaluate the traditional TTFS method in the same quantized-time domain as Otters++, while keeping the weights, architecture, and hyperparameters unchanged. In this setting, traditional quantized-time TTFS still requires digital time encoding, additional MAC-style operations, and extra weight accesses to evaluate the temporal term. With $T=15$ and the simple $T-t$ encoding, the traditional TTFS attention block consumes 19.09 mJ, which is 34.3\% higher than Otters++ under the same energy model. This comparison isolates the benefit of replacing digital temporal decoding with the device-native optical decay response.

% We can group TTFS encoding into two main classes: continuous-time TTFS and quantized-time TTFS. In other work, like TTFSFormer~\cite{zhaottfsformer}, the time domain is continuous. This means they pass the time-to-first-spike between layers using floating-point (FP32) values. This high-precision FP32 number helps keep the SNN accurate after the ANN–SNN conversion. However, comparing the energy of these continuous-time methods with our approach, which uses quantized time, is not fair. In TTFSFormer, their main math unit (e.g., the accumulator, ACC) is FP32 and costs about 0.9 pJ per operation, while our INT4 ACC only costs ~0.05 pJ. Also, the data movement cost grows with bit-width. This means continuous-time TTFS designs use much more memory and movement energy than low-bit designs.

% To make a fair comparison, we re-evaluated the continuous-time TTFS method in the same quantized time domain as Otters. We did this by quantizing the spike times while keeping the weights, network structure, and all other hyperparameters the same. This lets us clearly see the benefit of Otters' analog-based computing. Traditional quantized-time TTFS requires digital encoding, additional MAC operations (\(E_{\text{encoding}}{+}E_{\text{MAC}}\)), and extra weight accesses. With \(T{=}15\) using the simplest \(T{-}t\) encoding (4--4--4-bit ACC, \(0.0163\,\mathrm{pJ}\)), traditional TTFS attention block consumes 19.09mJ, \(34.3\%\) more energy than Otters++.

\subsection{Inference time analysis}

In Otters++, the inference latency is mainly determined by the duration of the optical decay function. Within this time window, the SNN neuron compares its accumulated membrane potential with the dynamic threshold at each time step $t_k$. If the membrane potential is larger than the threshold, the neuron fires. It then samples the value from the thin-film transistor and adds it to the next neuron’s membrane potential. Once the decay period finishes, the next layer is ready to start. Thus, in theory, the computation for one layer is completed within a single time window. Taking a transformer block as an example, the calculation requires roughly 8 optical cycles: 6 cycles for the Self-Attention module (projections of Q, K, V, their multiplication and output) and 2 cycles for the Feedforward module. 

In our current measurements, the transient optoelectronic response was recorded using a source-measure unit (SMU), whose minimum integration time is on the order of tens of microseconds. To ensure a window that can reliably acquire complete attenuation and fit parameters, we used a conservative sampling configuration, which resulted in a coding window of about 100 $\mu$s. This results in a total latency of ~0.8 ms for one block. Further, in theory, the window can be compressed by faster readout and higher light intensity.

  \section{Discussion and Future works}

To make SNNs more energy-efficient, Otters++ focuses on hardware-software co-design to optimize a core computing operation. We replace the temporal decay function and its following multiplication by sampling the natural decay of an oxide optoelectronic synapse. This shifts the computation from digital to analog for a more energy-efficient computing method. However, at the same time, many related works are also working toward better energy efficiency for SNNs, but from another direction, such as algorithmic and architectural improvements. These methods, including QKFormer, SSSA, and A$^2$OS$^2$A, optimize the spiking attention mechanism from $O(N^2)$ to linear complexity (linear-attention SNNs)~\cite{zhou2024qkformer,wang2025spiking,guo2025spiking}.These two directions are not in conflict. They are complementary and solve different parts of the problem. Future work includes designing a linear attention mechanism based on the Otters++ spiking neuron to make energy use even more efficient. We further discuss the hardware scale. The In$_2$O$_3$ optoelectronic synapse in our prototype has an effective area of about 0.012 mm$^2$ with a channel length of 30 $\mu$m. This size is mainly due to the precision limits of our fabrication equipment and does not represent the scaling limits of In$_2$O$_3$ technology. Since the goal of this paper is to validate the optoelectronic TTFS mechanism, we did not focus on device size optimization. However, recent work has demonstrated In$_2$O$_3$ transistors with channel lengths down to 40 nm~\cite{si2021high}, showing that the device area can be reduced by nearly three orders of magnitude. Additionally, several studies show that oxide devices can be integrated in 3D stacked layers, providing another path to further reduce the effective area per synapse ~\cite{tang2022monolithic,yuvaraja2024three,kwak2024monolithic}.

\section{Conclusion}
This paper presents Otters++, a physically grounded TTFS framework for energy-efficient spiking Transformer. By repurposing the natural signal decay of a custom-fabricated In$_2$O$_3$ optoelectronic synapse, Otters++ directly realizes the TTFS temporal term in device physics and avoids explicit digital decay evaluation. To make this mechanism trainable for Transformer models, we establish a layer-wise equivalence between Otters++ and an unsigned QNN, and develop a hybrid SNN-forward/QNN-backward training method with knowledge distillation and measured-variation-aware forward sampling. On GLUE, Otters++ reaches an average score of 84.17, improving over the original Otters baseline and outperforming prior SNN-based Transformer baselines. Under the same system-level energy model, Otters++ achieves the lowest per-layer energy among the compared models, with 1.84$\times$--5.68$\times$ energy reduction over prior spiking Transformer baselines. These results show that device-native TTFS computation can improve the trainability, robustness, and energy efficiency of spiking Transformers under realistic hardware effects.

%This paper introduces Otters++, a new paradigm for energy-efficient neuromorphic computing that challenges this digital-centric approach. Through a hardware-software co-design, we repurpose the natural signal decay of a custom-fabricated optoelectronic synapse, transforming this physical phenomenon into a computational method. This allows us to eliminate the costly evaluation of the decay function inherent in traditional TTFS, thereby fusing computation and memory into the physical process. To deploy this paradigm in complex architectures such as transformers, we developed a QNN-to-SNN conversion algorithm that circumvents the challenges of direct SNN training. The Otters++ model achieves state-of-the-art accuracy among SNNs on seven GLUE benchmark datasets, while simultaneously delivering a 1.77$\times$ improvement in energy efficiency over previous leading spiking models. By directly harnessing fundamental device physics for computation, this work demonstrates a new path to a more energy-efficient neuromorphic computing design. 

\section{Acknowledgement}
This research is partially supported by the Ministry of Education, Singapore, under the Academic Research Fund Tier 1 (FY2024).

\bibliographystyle{IEEEtran}
\bibliography{ref}

\vspace{-15 mm} 
\begin{IEEEbiography}[{\includegraphics[width=1in,height=1.25in,clip,keepaspectratio]{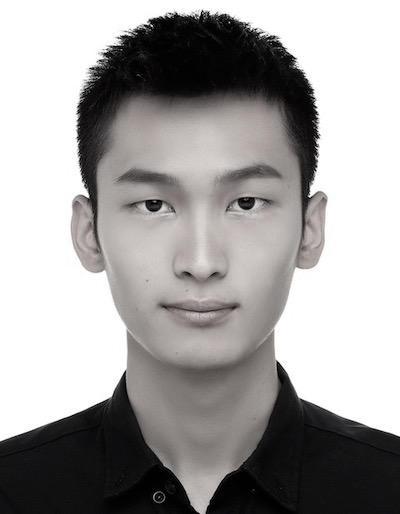}}]{Zhanglu Yan} received the BSc degree in computer science from Xian Jiaotong University, in 2019, and the MSc degree in artificial intelligence from National University of Singapore, in 2020. He is working toward the PhD degree in the School of Computing, National University of Singapore. His research focuses on neuromorphic computing and spiking neural networks.
\end{IEEEbiography}

\begin{IEEEbiography}
[{\includegraphics[width=1in,height=1.25in,clip,keepaspectratio]{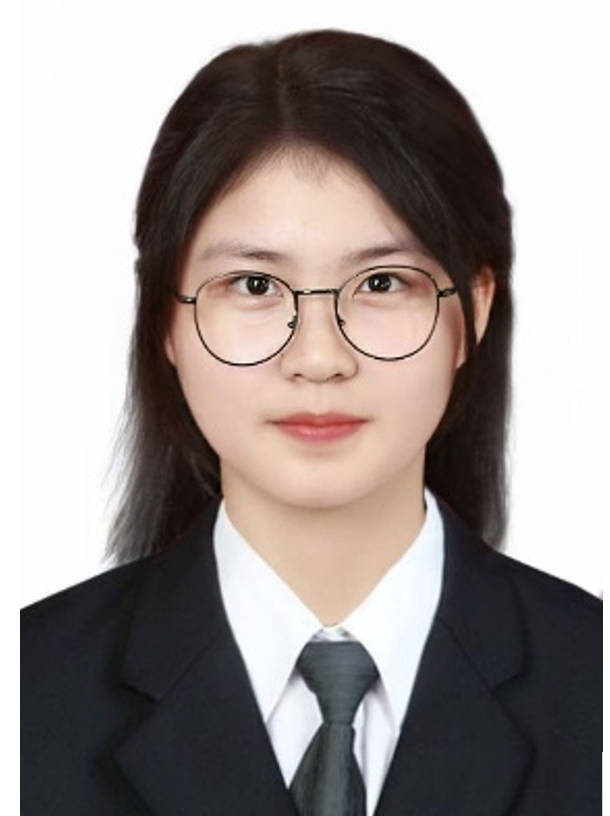}}]{Mao jiayi} received the BEng degree in electronic information engineering from Zhejiang Normal University in 2023. She is working toward the PhD degree in electronic science and technology at Westlake University. Her research focuses on oxide thin-film transistor-based artificial neuron devices for neuromorphic computing and sensing applications.
\end{IEEEbiography}

\begin{IEEEbiography}[{\includegraphics[width=1in,height=1.25in,clip,keepaspectratio]{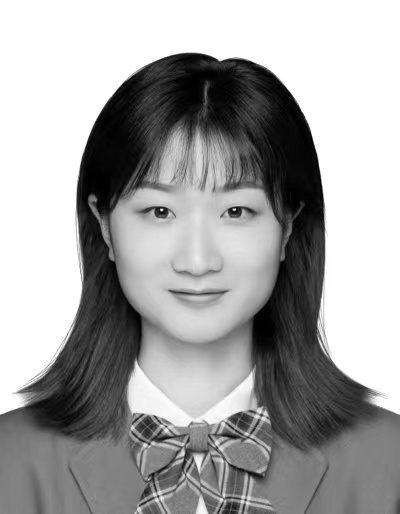}}]{Kaiwen Tang} received the BSc degree in computer science from Xi'an Jiaotong University in 2022. She is working toward the PhD degree in the School of Computing, National University of Singapore. Her research focuses on spiking neural networks and efficient language models.
\end{IEEEbiography}

\begin{IEEEbiography}
[{\includegraphics[width=1in,height=1.25in,clip,keepaspectratio]{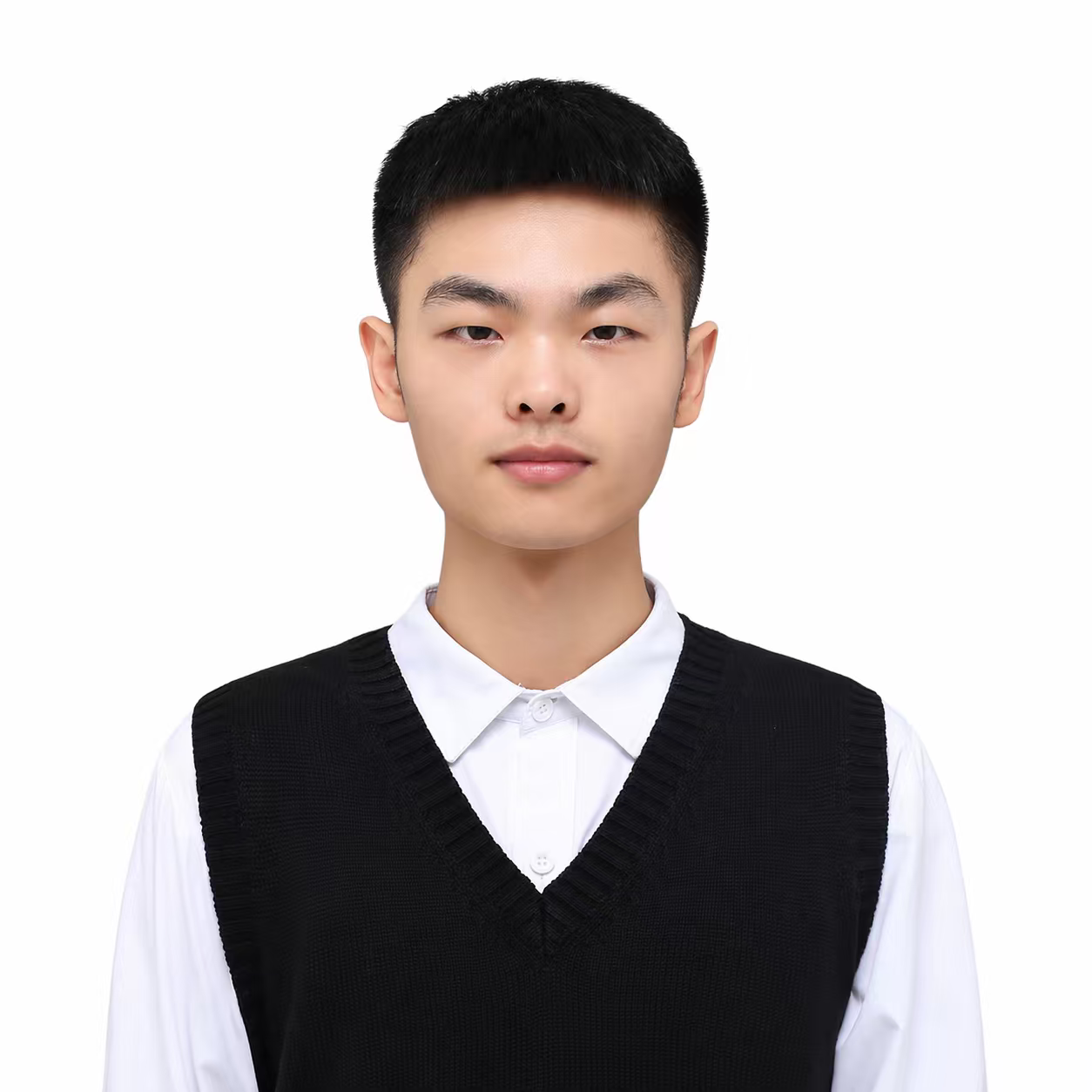}}]{Li fanfan} received the B.E. degree in inorganic non-metallic materials from Xi'an University of Science and Technology in 2019, and the M.S. degree in materials science and engineering from Xidian University in 2022. He is currently pursuing the Ph.D. degree in materials science and engineering at Zhejiang University. His research focuses on neuromorphic sensing systems and memristor-based artificial neurons.
\end{IEEEbiography}

\begin{IEEEbiography}
[{\includegraphics[width=1in,height=1.25in,clip,keepaspectratio]{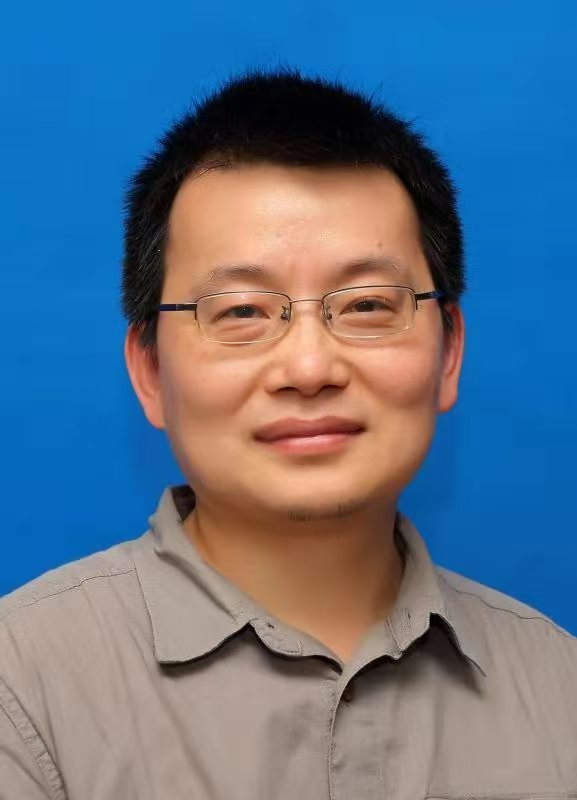}}]{Gang Pan} is a professor of the College of Computer Science and Technology, and vice-director of State Key Lab of CAD\&CG, Zhejiang University. He received the B.Eng. and Ph.D. degrees from Zhejiang University in 1998 and 2004 respectively. From 2007 to 2008, he was a visiting scholar at the University of California, Los Angeles. His interests include artificial intelligence, brain-inspired computing, brain-machine interfaces, and pervasive computing. He has co-authored more than 100 refereed papers, and has 49 patents granted. He serves as Associate Editors of IEEE Systems Journal, IEEE Transactions on Neural Networks and Learning Systems, IEEE Transactions Cybernetics, Pervasive and Mobile Computing, and IEEE Transactions on Cognitive and Developmental Systems.
\end{IEEEbiography}

\begin{IEEEbiography}
[{\includegraphics[width=1in,height=1.25in,clip,keepaspectratio]{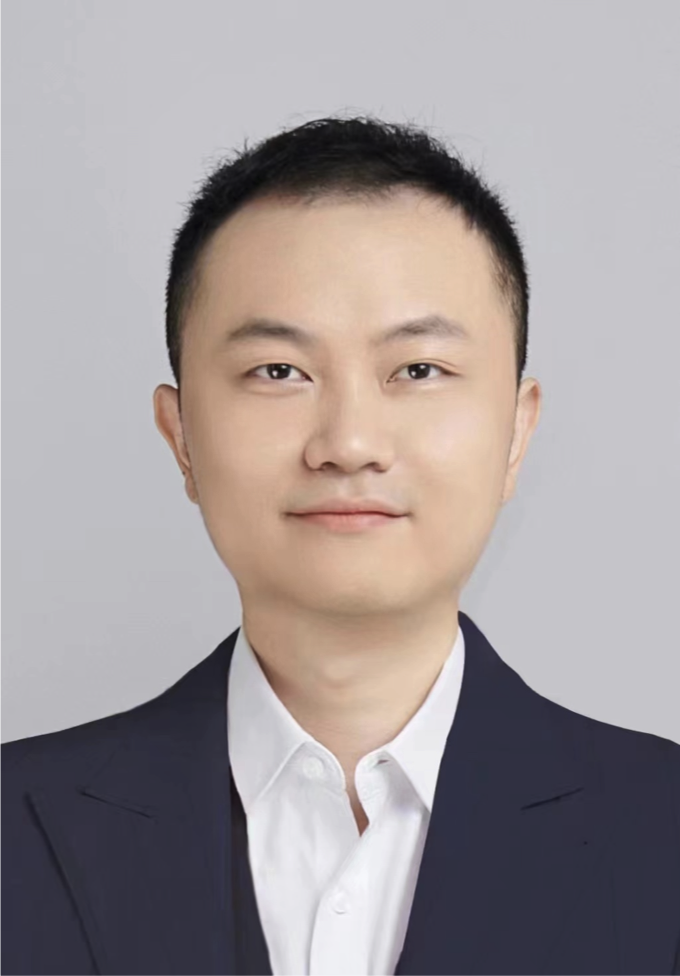}}]{Luo Tao} received his bachelor’s degree from the Harbin Institute of Technology, Harbin, China, in 2010, his master’s degree from the University of Electronic Science and Technology of China, Chengdu, China, in 2013, and his Ph.D. degree from the School of Computer Science and Engineering, Nanyang Technological University, Singapore, in
2018. He is currently a senior research scientist with the Institute of High Performance Computing (IHPC), Agency for Science, Technology and Research, Singapore (A*STAR), Singapore. His current research interests include high-performance computing, machine learning,
computer architecture, hardware–software co-exploration, quantum comput-
ing, efficient AI and its application.
\end{IEEEbiography}

\begin{IEEEbiography}
[{\includegraphics[width=1in,height=1.25in,clip,keepaspectratio]{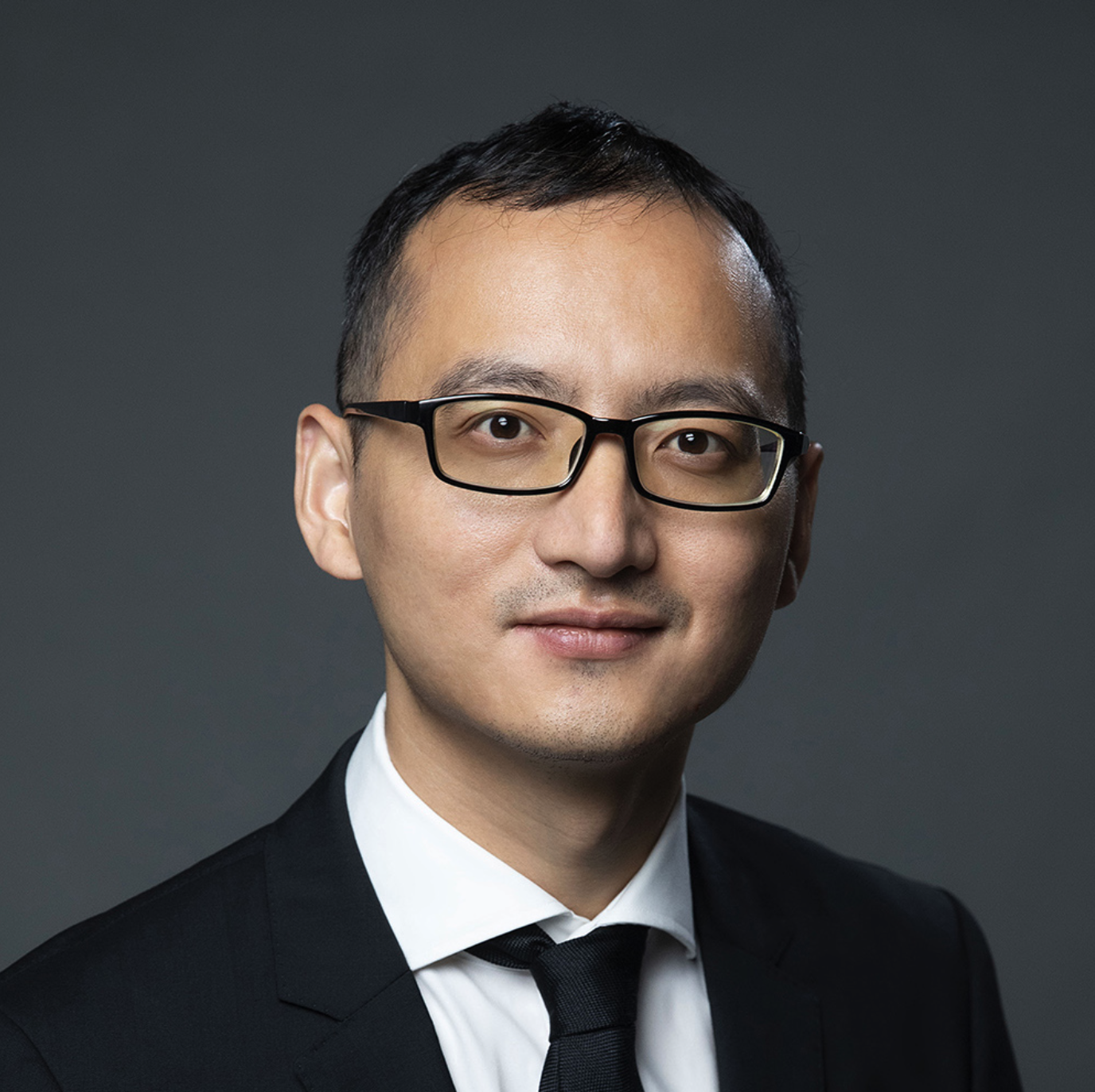}}]{Zhu bowen} received his BS degree in chemistry from Jilin University (China) in 2010. He obtained his PhD degree in materials science in 2016 from Nanyang Technological University, Singapore. After his postdoctoral fellow training at UCLA, he moved to Monash University in 2017, Australia, as a Discovery Early Career Researcher Award (DECRA) fellow. He joined School of Engineering, Westlake University, China, as an independent Principal Investigator (PI) in August 2019. His research is focused on developing high-performance flexible metal oxide thin-film transistors (TFTs) to construct active-matrix sensing arrays by integrating TFT backplane with physical sensors, biosensors, and displays. Meanwhile, he is also interested in developing artificial spiking sensory systems by integrating Mott memristor neurons into sensors. He has published over 80 peer-reviewed papers on leading journals of materials and electronics research with a total citation of over 8000 with a H-index of 40.
\end{IEEEbiography}

\begin{IEEEbiography}
[{\includegraphics[width=1in,height=1.25in,clip,keepaspectratio]{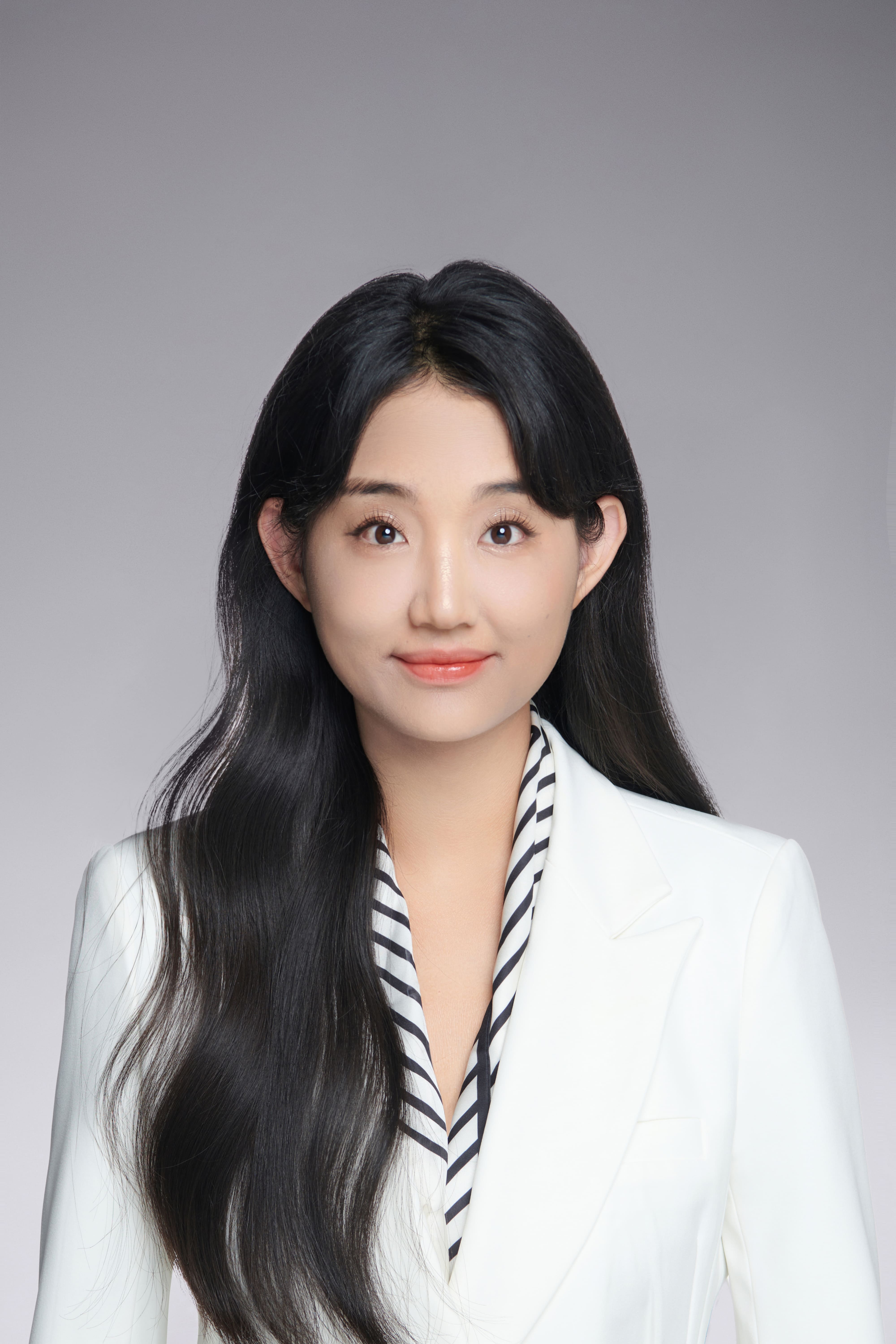}}]{Qianhui Liu}
received the B.S. and Ph.D. degrees in the College of Computer Science and Technology from Zhejiang University in 2016 and 2021 respectively. She is currently an Associate Professor in the School of Artificial Intelligence, Shandong University. From 2022 to 2025, she was a research fellow at the Department of Electrical and Computer Engineering, National University of Singapore, Singapore. Her research interests include neuromorphic computing, spiking neural networks, and audio-visual speech recognition.
\end{IEEEbiography}
\vspace{-155 mm} 
\begin{IEEEbiography}
[{\includegraphics[width=1in,height=1.25in,clip,keepaspectratio]{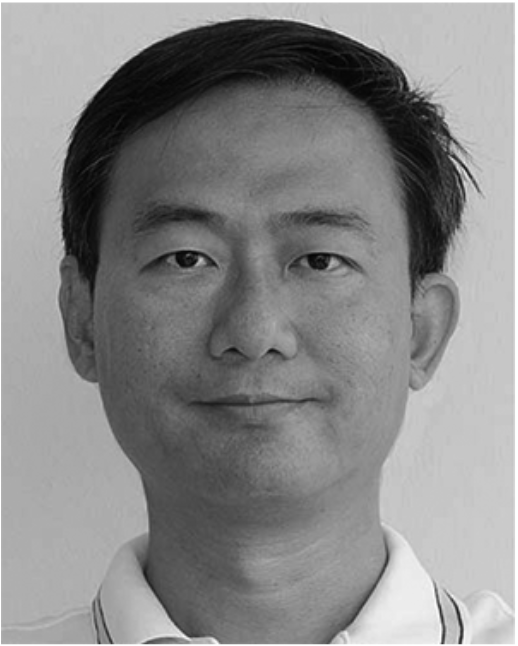}}]{Wong Weng-Fai} received the BSc degree from the National University of Singapore, in 1988, and the DrEngSc degree from the University of Tsukuba, Japan, in 1993. He is currently an associate professor with the Department of Computer Science, National University of Singapore. His research interests include computer architecture, compilers, and high-performance computing. He is a senior member of the IEEE.
\end{IEEEbiography}

\end{document}